\documentclass{article}

\PassOptionsToPackage{numbers, compress}{natbib}

\usepackage[preprint]{neurips_2026}

\usepackage[utf8]{inputenc} 
\usepackage[T1]{fontenc}    
\usepackage{hyperref}       
\usepackage{url}            
\usepackage{booktabs}       
\usepackage{amsfonts}       
\usepackage{nicefrac}       
\usepackage{microtype}      
\usepackage{xcolor}         
\usepackage{microtype}
\usepackage{graphicx}
\usepackage{subcaption}
\usepackage{booktabs}
\usepackage{multirow}
\usepackage{pifont}
\usepackage{amsmath}
\usepackage{amssymb}
\usepackage{mathtools}
\usepackage{amsthm}
\usepackage{tabularx}
\usepackage{makecell}
\usepackage{enumitem}
\usepackage{algorithm}
\usepackage{algorithmic}
\usepackage{cleveref}
\usepackage{wrapfig}
\usepackage{float} 
\usepackage{url}

\usepackage{xspace}

\newcommand{\Wikipedia}{\text{Wikipedia}\xspace}
\newcommand{\Reddit}{\text{Reddit}\xspace}
\newcommand{\Mooc}{\text{MOOC}\xspace}
\newcommand{\Enron}{\text{Enron}\xspace}
\newcommand{\UCI}{\text{UCI}\xspace}
\newcommand{\USLegis}{\text{US Legis.}\xspace}
\newcommand{\UNVote}{\text{UN Vote}\xspace}
\newcommand{\UNTrade}{\text{UN Trade}\xspace}
\newcommand{\CanParl}{\text{Can. Parl.}\xspace}

\newcommand{\DiffDyG}{\text{DiffDyG}\xspace}
\newcommand{\RepeatMixer}{\text{RepeatMixer}\xspace}
\newcommand{\TIDFormer}{\text{TIDFormer}\xspace}
\newcommand{\DyGFormer}{\text{DyGFormer}\xspace}
\newcommand{\CNEN}{\text{CNEN}\xspace}
\newcommand{\TGAT}{\text{TGAT}\xspace}
\newcommand{\TGN}{\text{TGN}\xspace}
\newcommand{\TCL}{\text{TCL}\xspace}
\newcommand{\CAWN}{\text{CAWN}\xspace}
\newcommand{\EdgeBank}{\text{EdgeBank}\xspace}
\newcommand{\GraphMixer}{\text{GraphMixer}\xspace}

\newcommand{\cmark}{\ding{51}}
\newcommand{\xmark}{\ding{55}}

\crefformat{section}{\S#2#1#3} 
\crefformat{subsection}{\S#2#1#3}

\usepackage{amsmath,amsfonts,bm}

\def\eqref#1{equation~\ref{#1}}

\def\1{\bm{1}}

\def\rve{{\mathbf{e}}}

\def\rvu{{\mathbf{i}}}

\def\rvu{{\mathbf{u}}}

\DeclareMathAlphabet{\mathsfit}{\encodingdefault}{\sfdefault}{m}{sl}
\SetMathAlphabet{\mathsfit}{bold}{\encodingdefault}{\sfdefault}{bx}{n}

\def\gG{{\mathcal{G}}}

\title{Attention Dispersion in Dynamic Graph Transformers: Diagnosis and a Transferable Fix}

\author{%
  Jinhao Zhang$^1$ \quad Kangfei Zhao$^1$ \quad Qiuhao Zeng$^2$ \quad
  Long-Kai Huang$^3$\stepcounter{footnote}\thanks{Correspondence to: Long-Kai Huang \texttt{<longkai@hkbu.edu.hk>}} \\
  $^1$Beijing Institute of Technology \\
  $^2$University of Toronto \\
  $^3$Hong Kong Baptist University \\
}

\begin{document}

\maketitle

\begin{abstract}
Transformer-based architectures have become the dominant paradigm for Continuous-Time Dynamic Graph (CTDG) learning, yet their performance remains limited on temporally shifted datasets. 
In this work, we identify \emph{attention dispersion} as a shared failure mode of dynamic graph Transformers under temporal distribution shift. 
Through controlled ablation contrasting structurally and temporally distinguished historical neighbors against random ones, we show that prediction depends on a class of \emph{critical nodes} that carry consistently more predictive signal than arbitrary neighbors. However, existing Transformers fail to focus on these nodes even when they are present in the input, as temporal shift weakens attention contrast and produces overly dispersed attention distributions. 
This diagnosis suggests a simple and transferable fix: replace standard attention with differential attention, which suppresses common-mode attention and amplifies distinctive token-level signals. 
When added to three representative CTDG Transformer baselines, differential attention consistently improves performance, with gains concentrated on high-shift datasets. 
Attention-level measurements further confirm the mechanism, showing reduced attention entropy and increased attention mass on critical nodes. 
Building on these findings, we introduce \DiffDyG, a reference implementation combining differential attention with standard input encodings. 
Across 9 benchmarks and three negative sampling protocols, \DiffDyG achieves SOTA performance, with especially large gains on the most shifted datasets.
\end{abstract}

\section{Introduction}\label{sec:intro}

Continuous-Time Dynamic Graphs (CTDGs) provide a natural framework for modeling temporal interactions in evolving systems, including social networks, e-commerce platforms, communication infrastructures, and biological processes~\citep{DBLP:conf/nips/Huang0W0ZXCV22,kumar2019predicting,DBLP:conf/kdd/HuangHRR20}. 
Recent CTDG models increasingly adopt Transformer-based architectures~\citep{DBLP:conf/nips/VaswaniSPUJGKP17}, which use self-attention to process variable-length historical interaction sequences. 
A line of recent work~\citep{Wang2021TCL,yu2023towards,Peng2025TIDFormer} has improved how dynamic graph structures are converted into Transformer-compatible sequences, for example through neighbor co-occurrence encoding and temporal patching. 
These designs allow Transformers to model longer histories and have made them a dominant paradigm on standard CTDG benchmarks.

Despite these advances, existing Transformer-based models plateau on the same several datasets. Across the 9 standard CTDG benchmarks we examine, the strongest existing transformer reaches only $71.1\%$ AP on \USLegis, $66.5\%$ on \UNTrade, and $69.0\%$ on \UNVote, with little improvement despite years of architectural innovation in sequence construction and feature design. The pattern is consistent across architectures rather than specific to any one method, and the affected datasets correlate with high temporal distribution shift between training and test windows. This suggests a shared failure mode under temporal distribution shift. Existing approaches, all of which target sequence construction or feature design, do not fix it.

We trace this limitation to attention allocation. 
Through controlled ablation, we identify a class of historical neighbors that we call \emph{critical nodes}. 
These nodes either occupy central positions in the local interaction structure or participate in temporally stable relationships. Masking critical nodes causes a much larger performance drop on shifted datasets than masking the same number of randomly selected neighbors, showing that these nodes carry disproportionately important predictive information. However, existing Transformer-based models still perform poorly on shifted datasets even when all critical nodes are present in the input. Thus, the issue is not simply information availability. It is also unlikely to be a pure capacity issue, since the same architectures perform well on less-shifted datasets. Instead, the failure lies in how attention distributes probability mass over historical tokens. Under temporal shift, the attention score provides weaker relative separation among tokens, and the subsequent softmax produces a flatter attention distribution. As a result, attention becomes more dispersed, and the model fails to concentrate on the critical historical signals that remain present in the input.

This diagnosis leads to two testable predictions. 
First, if attention allocation is the shared bottleneck, then replacing only the attention mechanism in existing Transformers should improve their robustness under temporal shift while leaving their sequence construction and other components unchanged. 
Second, the improved attention mechanism should produce less dispersed attention distributions and assign more mass to critical nodes. 
We test these predictions using differential attention~\citep{ye2024differential}, originally proposed for language modeling, which subtracts two softmax attention maps computed over the same input, thereby suppressing dispersed common-mode attention and amplifying distinctive token-level signals.

Both predictions are supported empirically. 
When added as a plug-in module to three architecturally different Transformer baselines, namely \DyGFormer, \TIDFormer, and \TCL, differential attention improves their average AP by $+8.3$, $+5.7$, and $+2.9$, respectively. 
The gains are concentrated on high-shift datasets, where improvements range from $+11$ to $+26$ AP. 
Direct attention-level measurements further support the proposed mechanism. 
Differential attention consistently reduces attention entropy and increases the proportion of attention mass assigned to critical nodes, with the strongest effects appearing on the datasets where temporal distribution shift is most severe. 
Together, the failure-mode analysis, the cross-architecture performance recovery, and the attention-level measurements point to the same conclusion: attention allocation is a shared bottleneck of CTDG Transformers under temporal distribution shift, and differential attention provides a targeted remedy.

Building on this finding, we introduce \DiffDyG, a CTDG Transformer that combines differential attention with standard input-construction components, including RoPE, neighbor co-occurrence encoding, and spatial distance encoding. Across 9 benchmarks, \DiffDyG achieves the strongest overall performance. 
On the three most challenging shifted datasets, it improves AP from $71.1$ to $87.5$ on \USLegis, from $66.5$ to $99.0$ on \UNTrade, and from $69.0$ to $88.8$ on \UNVote. These results show that a targeted change to attention allocation can close performance gaps that prior sequence-construction improvements have not resolved.

Our contribution is to identify where Transformer-based CTDG models fail under temporal distribution shift, show that this failure is shared across architectures, validate a targeted attention-level fix, and verify the mechanism through direct measurements of attention behavior. Contributions are summarized as:
\begin{itemize}[noitemsep,topsep=0pt]
    \item[1)] We diagnose attention dispersion as a shared failure mode of Transformer-based CTDG models under temporal distribution shift. Through controlled ablations on critical and random historical neighbors, we show that key predictive signals are present in the input but are not properly attended to under shift.
    \item[2)] We validate this diagnosis across architectures and mechanisms. Adding differential attention to three representative Transformer baselines yields consistent gains, especially on high-shift datasets, while direct attention-level measurements show reduced attention entropy and increased attention mass on critical nodes.
    \item[3)] We introduce \DiffDyG, a reference implementation that combines differential attention with standard input-construction components and achieves new SOTA performance on all 9 benchmarks under three negative sampling protocols.
\end{itemize}
\section{Related works}
\textbf{Dynamic graph learning.}
Existing dynamic graph modeling approaches broadly fall into two categories: discrete-time modeling~\cite{cong2021dynamic, sankar2020dysat, you2022roland} and continuous-time modeling~\cite{cong2023we, souza2022provably, wang2021inductive}. Discrete-time dynamic graph (DTDG) methods represent a dynamic graph as a sequence of snapshots. Static graph encoders are applied to each snapshot and sequential modules capture the time-series dynamics~\cite{hu2021time, Pareja2020EvolveGCN}.
In contrast, continuous-time dynamic graph (CTDG) methods represent a graph as a stream of timestamped interactions, enabling fine-grained temporal reasoning.
CTDG models include temporal Point Process (TPP)-based models~\cite{han2019graph, trivedi2017know, zhao2023time}, random-walk based models~\cite{li2023zebra, lu2024improving,  wang2021inductive}, memory-based models~\cite{rossi2020temporal, wang2021apan, su2024pres}, and specialized temporal message-passing networks~\cite{kumar2019predicting, yu2023towards, wu2024feasibility}, where Transformer-based~\cite{DBLP:conf/nips/VaswaniSPUJGKP17} architectures have achieved strong performance in CTDG learning. 
Specifically, TCL~\cite{Wang2021TCL} combines a graph Transformer with contrastive learning to master temporal and topological information in dynamic graphs. 
DyGFormer~\cite{yu2023towards} takes a neighbor co-occurrence encoding as the input of vanilla Transformers. 
SimpleDyG~\cite{wu2024feasibility} proposes an ego-graph tokenization scheme along the temporal dimension, and TIDFormer~\cite{Peng2025TIDFormer} designs calendar-based temporal encodings for mixed-granularity interaction modeling. 
Despite these research efforts, most existing methods neglect the temporal shift problem underlying dynamic graphs.


\section{Diagnosing the Failure Mode of Dynamic Graph Transformers}\label{sec:2}

In this section, we investigate why transformer-based CTDG models systematically degrade under temporal distribution shift. We show that performance degradation is strongly associated with temporal distribution shift and that the same datasets are difficult for several Transformer architectures (\cref{sec:2_measure_gap}). To localize the cause, we identify a class of historical neighbors carrying additional predictive signal, which we define as \emph{critical nodes}, and show through controlled ablation that the failure on shifted data is not one of information availability or model capacity but of attention allocation (\cref{sec:2_critical_node}). We first formalize the problem and setup.

\textbf{Continuous-Time Dynamic Graph (CTDG).} A CTDG is a chronologically ordered sequence of temporal interactions $\mathcal{G} = \{(u_1, v_1, t_1), \cdots, (u_n, v_n, t_n)\}$, where $0 \le t_1 \le \cdots \le t_n$. Each tuple $(u_i, v_i, t_i)$ represents a directed interaction at time $t_i$ between nodes $u_i, v_i \in \mathcal{N}$. Nodes and interactions may have associated features $\rvu\in\mathbb{R}^{d_N}$ and $\rve\in\mathbb{R}^{d_E}$..

\textbf{Problem definition.} Given a source $u$, destination $v$, timestamp $t$, and the history $\{(u', v', t') \mid t' < t\}$, dynamic link prediction estimates the likelihood of an interaction between $u$ and $v$ at time $t$.

\textbf{Datasets and evaluation.} We use 9 standard CTDG benchmarks~\cite{poursafaei2022towards}: \Wikipedia, \Reddit, \Mooc, \Enron, \UCI, \CanParl, \USLegis, \UNTrade, and \UNVote. Following~\cite{yu2023towards}, we use chronological 70\%/15\%/15\% train/validation/test splits. Unless otherwise specified, we use random negative sampling strategy and report Average Precision (AP) in transductive setting. For diagnosis, we analyze 3 Transformer-based methods with different sequence construction designs: \DyGFormer~\cite{yu2023towards}, \TIDFormer~\cite{Peng2025TIDFormer}, and \TCL~\cite{Wang2021TCL}. Experimental details are in Appendix~\ref{app:exp_details}.

\subsection{Temporal Shift Exposes a Shared Failure Pattern}\label{sec:2_measure_gap}

We quantify temporal distribution shift between the training and test windows using Maximum Mean Discrepancy (MMD)~\cite{gretton2012kernel}. MMD is computed on features extracted from the two windows using trained DyGFormer~\cite{yu2023towards}. 

\begin{table*}[!t]
\caption{\small Temporal distribution gap measured by Maximum Mean Discrepancy (MMD) across datasets.}
\label{tab:mmd_datasets}
\vspace{-1ex}
\centering
\scriptsize
\setlength{\tabcolsep}{6pt}
\resizebox{0.98\textwidth}{!}{%
\begin{tabularx}{\textwidth}{l*{9}{>{\centering\arraybackslash}X}}
\toprule
Metric & \Wikipedia & \Reddit & \UCI & \Enron & \Mooc & \CanParl & \USLegis & \UNTrade & \UNVote \\
\midrule
MMD & 0.271 & 0.257 & 0.340 & 0.457 & 0.355 & 0.372 & 0.697 & 0.476 & 0.752 \\
\bottomrule
\end{tabularx}%
}
\vspace{-2ex}
\end{table*}

\begin{figure*}[!t]
  \centering
  \begin{subfigure}[b]{0.32\linewidth}
    \centering
    \includegraphics[width=\linewidth]{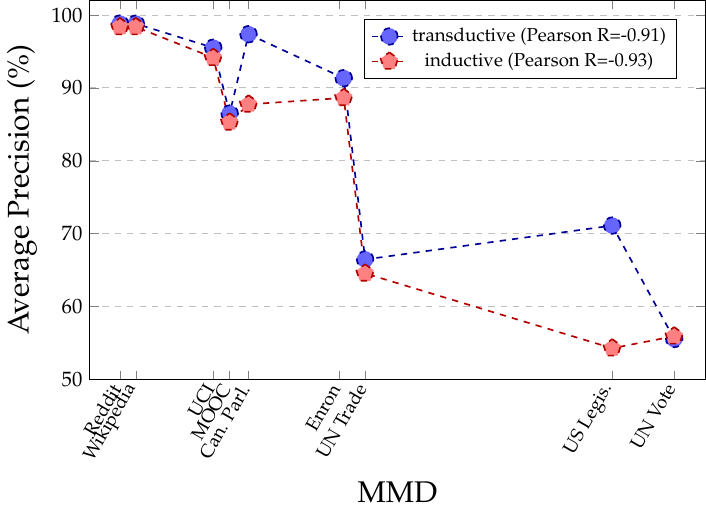}
    \caption{\small \DyGFormer}\label{fig:corr:dygformer}
  \end{subfigure}\hfill
  \begin{subfigure}[b]{0.32\linewidth}
    \centering
    \includegraphics[width=\linewidth]{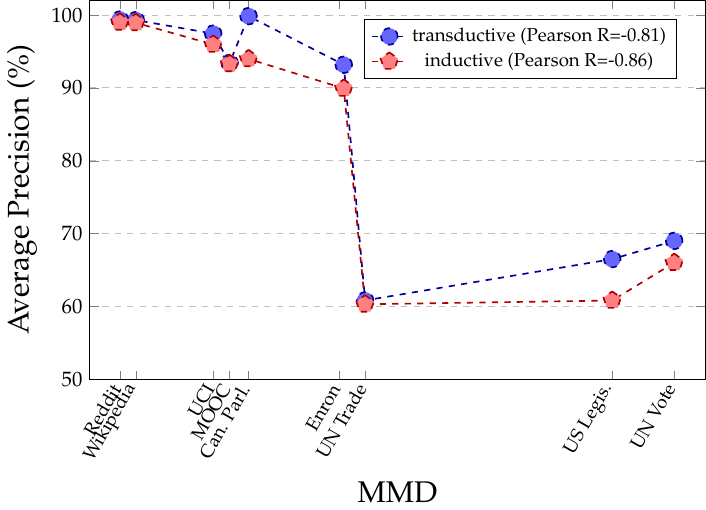}
    \caption{\small \TIDFormer}\label{fig:corr:tidformer}
  \end{subfigure}\hfill
  \begin{subfigure}[b]{0.32\linewidth}
    \centering
    \includegraphics[width=\linewidth]{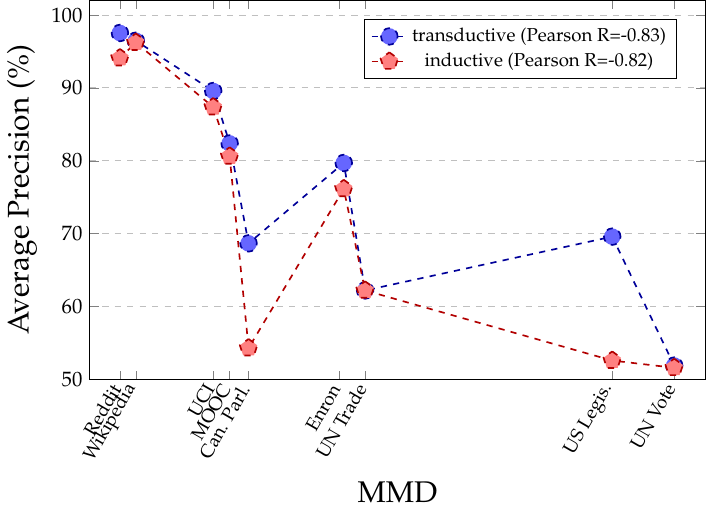}
    \caption{\small \TCL}\label{fig:corr:tcl}
  \end{subfigure}
  \caption{\small Performance on 9 datasets with varying MMD. Legend reports Pearson's R between AP and MMD.}
  \label{fig:mmd:performance_vs_shift}
\vspace{-4ex}
\end{figure*}


Tab.~\ref{tab:mmd_datasets} shows that the benchmarks differ substantially in temporal shift. \Wikipedia and \Reddit have low shift, with MMD below $0.30$, whereas \USLegis and \UNVote exceed $0.69$. \UNTrade has the third-largest MMD. Fig.~\ref{fig:mmd:performance_vs_shift} shows a strong negative correlation between MMD and AP for all 3 Transformer baselines in both transductive and inductive settings, with Pearson's $R$ ranging from $-0.81$ to $-0.93$.

The consistency across architectures is important. \DyGFormer, \TIDFormer, and \TCL construct historical sequences in different ways, yet they degrade on the same high-shift datasets. In particular, the best existing Transformer reaches only $71.1\%$ AP on \USLegis, $66.5\%$ on \UNTrade, and $69.0\%$ on \UNVote. This pattern suggests that the bottleneck is unlikely to be tied only to a specific sequence construction strategy. Instead, it points to a component shared by these models.

\subsection{Critical neighbors as a Diagnostic Probe}\label{sec:2_critical_node}


To localize the cause of the shared failure, we examine which historical interactions carry the predictive signal needed for link prediction, whether existing models exploit them, and what changes between low-shift and high-shift conditions.

\begin{wrapfigure}{r}{0.35\textwidth}
    \centering
    \vspace{-6ex}
    \includegraphics[width=\linewidth]{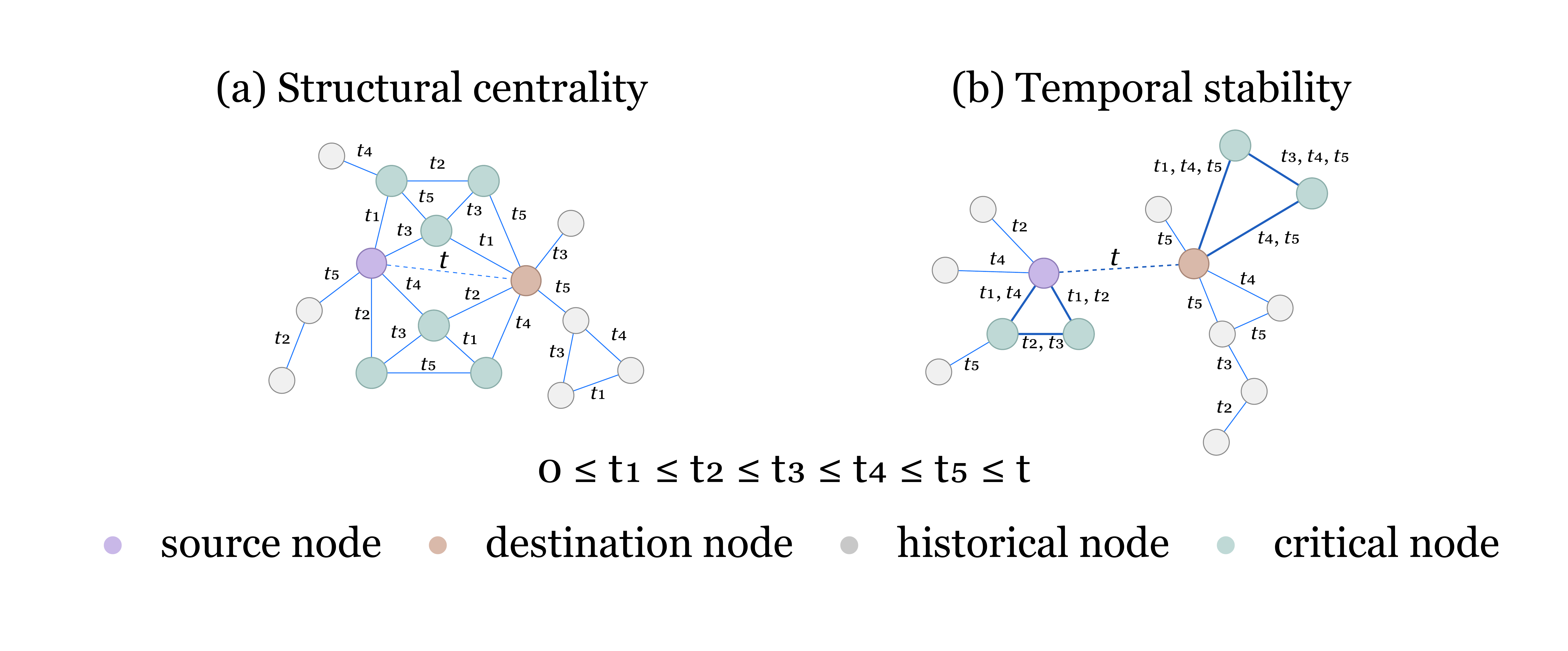}
    \vspace{-4ex}
    \caption{\small Illustration of the two criteria for critical nodes. The purple and brown nodes denote the source and destination, gray nodes denote ordinary historical neighbors, and green nodes denote critical nodes. The dashed edge is the query interaction at time $t$, and solid edges are historical interactions before $t$.}
    \vspace{-4ex}
    \label{fig:critical_node_illustration}
\end{wrapfigure}
\textbf{Critical nodes.} For a test query $(u,v,t)$, let $\mathcal{C}(u,v,t) = (\mathcal{H}(u,t) \cup \mathcal{H}(v,t)) \setminus \{u,v\}$ denotes the combined historical neighborhood of the source and destination before time $t$, where $\mathcal{H}(u,t)$ denotes the historical 1-hop neighborhoods of $u$. Not all neighbors in $\mathcal{C}$ are equally informative. We define a node $w\in\mathcal{C}(u,v,t)$ as a \emph{critical node} if it satisfies at least one of two properties: \textbf{1) Structural centrality:} $w$ is connected to multiple other nodes in $\mathcal{C}(u,v,t)$ and therefore reflects local community structure; \textbf{2) Temporal stability:} $w$ has interacted more than once with another node in $\mathcal{C}(u,v,t)$, and both nodes have multiple historical interactions with the source or destination, thereby reflecting a stable local relationship anchored to the query pair. The critical nodes are illustrated in Fig.~\ref{fig:critical_node_illustration}.

We denote the resulting set by $\mathcal{K}(u,v,t)$. This definition is not intended as the only possible definition of important neighbors. It is used as a diagnostic probe: if these nodes matter more than randomly selected neighbors of equal count, then they provide a useful way to test whether attention focuses on predictive historical signals.

\begin{figure}
    \centering
    \includegraphics[width=0.9\linewidth]{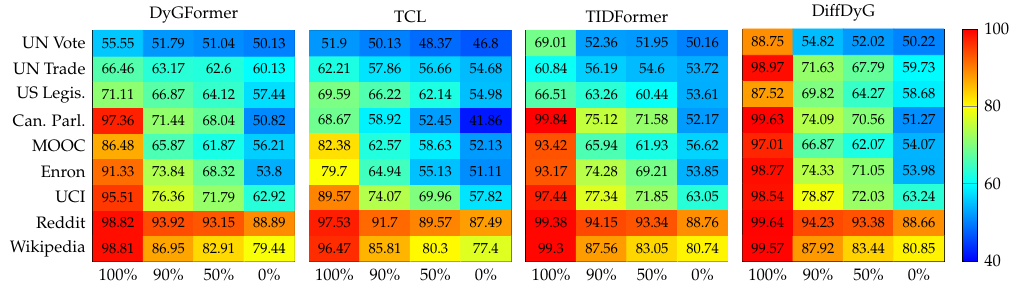}
    \caption{\small Critical-node ablation. The x-axis shows the retention ratio of critical nodes: 100\% keeps all critical nodes, while 0\% masks all critical nodes. Non-critical nodes are kept unchanged.}
    \label{fig:critical_nodes_ablation}
    \vspace{-2ex}
\end{figure}

\textbf{Critical-node ablation.}
For each test query $(u,v,t)$, we mask tokens corresponding to $\mathcal{K}(u,v,t)$ at different retention levels and evaluate AP. At a retention ratio of $r\%$, only $r\%$ of critical nodes are kept, while the remaining critical nodes are masked. All non-critical historical tokens are preserved. Fig.~\ref{fig:critical_nodes_ablation} reports the results across all nine datasets.

The results show that critical nodes carry substantial predictive signal. On low-shift datasets such as \Wikipedia and \Reddit, performance remains relatively high even when many critical nodes are removed, suggesting that useful signals are more redundant. On medium-shift datasets such as \UCI, \Enron, \Mooc, and \CanParl, models perform well when critical nodes are retained, but performance drops sharply as these nodes are removed. On high-shift datasets such as \USLegis, \UNTrade, and \UNVote, existing Transformer baselines already perform poorly even with all critical nodes available, and their performance further decreases after critical-node masking. This indicates that critical nodes are important, but their presence alone is insufficient under severe temporal shift.

\begin{figure}
    \centering
    \includegraphics[width=0.9\linewidth]{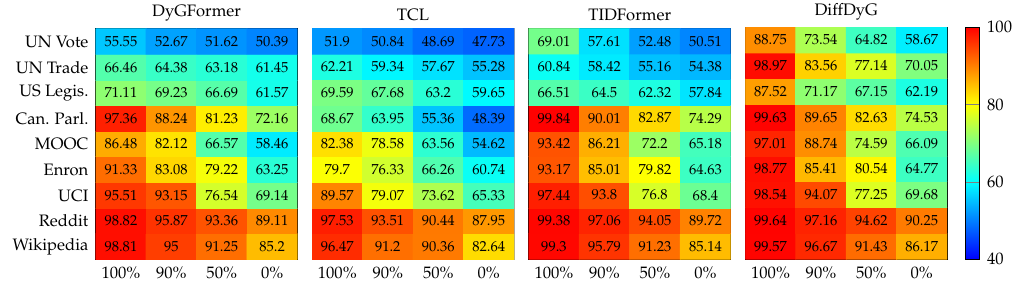}
    \caption{\small Random-node ablation. For each retention ratio, we mask the same number of randomly selected historical nodes as in the corresponding critical-node ablation. Thus, 0\% masks a random set with the same cardinality as the full critical-node set, rather than masking all historical nodes.}
    \label{fig:random_nodes_ablation}
    \vspace{-2ex}
\end{figure}

\textbf{Random-neighbor control.}
To test whether the above degradation is simply caused by removing historical tokens, we repeat the experiment with random masking. For each retention level, we mask the same number of randomly selected historical nodes as in the critical-node ablation. The random set is sampled from the full historical neighborhood and may include critical nodes by chance. Therefore, this control asks whether the structurally and temporally defined nodes in $\mathcal{K}(u,v,t)$ are more informative than an equally sized random subset. The comparison in Fig.~\ref{fig:random_nodes_ablation} confirms this. 

Across models and datasets, masking critical nodes is generally more damaging than masking the same number of random nodes. For example, on \USLegis with \DyGFormer, masking all critical nodes drops AP from $71.1\%$ to $57.4\%$, whereas random masking with the same cardinality drops AP only to $61.6\%$. Similar gaps appear on \UNTrade and \UNVote, and the pattern is also visible for \DiffDyG, where critical-node masking causes larger degradation than random masking under the same mask size. Since random masking can remove critical nodes by chance, this gap gives a conservative estimate of the additional predictive signal carried by critical nodes.


\textbf{Implication.} These ablations separate information availability from information use. Critical nodes are present in the input across all datasets. On medium-shift datasets, the same architectures can use them effectively. On high-shift datasets, the models still fail even when the critical nodes remain in the input. Thus, the main bottleneck is not simply missing information or insufficient capacity. Rather, it is the model's ability to assign enough relative weight to predictive historical tokens among many less informative ones. Since the same failure appears across different Transformer architectures, the evidence points to the shared attention mechanism.

\subsection{Hypothesis: Attention Dispersion under Shift}\label{sec:2_bottleneck}

We hypothesize that temporal distribution shift weakens the contrast produced by standard softmax attention. Standard attention computes $\mathrm{softmax}(QK^\top/\sqrt{d})V,$
where the softmax distributes probability mass over all historical tokens. When test-time interaction patterns differ from training-time patterns, the relative gaps in $QK^\top$ can become smaller. After softmax, smaller score gaps lead to flatter attention distributions. The model may still assign slightly higher mass to useful tokens, but the contrast is too weak for downstream prediction to reliably extract their signal.

We refer to this failure mode as \emph{attention dispersion}. The hypothesis makes two testable predictions. First, if attention allocation is the shared bottleneck, then replacing only the attention module in existing CTDG Transformers should improve performance on shifted datasets while leaving their sequence construction unchanged. Second, a successful replacement should produce less dispersed attention distributions and assign more mass to critical nodes. \cref{sec:validation} validates the diagnosis by testing both predictions directly.
\section{Validating the Diagnosis}\label{sec:validation}


\paragraph{Differential attention as an intervention.} We test both predictions in \cref{sec:2_bottleneck} using differential attention as a targeted intervention. Differential attention~\cite{ye2024differential} is suitable for this test because it changes the attention computation while leaving the surrounding architecture unchanged. Given an input representation {\small $\mathbf{Z}\in\mathbb{R}^{N\times d_{\mathrm{in}}}$}, we form two query-key pairs and a shared value: {\small $[\mathbf{Q}_1;\mathbf{Q}_2]=\mathbf{Z}\mathbf{W}^{Q}$, $[\mathbf{K}_1;\mathbf{K}_2]=\mathbf{Z}\mathbf{W}^{K}$, $\mathbf{V}=\mathbf{Z}\mathbf{W}^{V}.$}
Let $\mathbf{A}_m=\mathrm{softmax}\!\left(\frac{\mathbf{Q}_m\mathbf{K}_m^\top}{\sqrt{d_{\mathrm{attn}}}}\right), m\in\{1,2\}.$ Diff attention computes
\begin{equation}
\label{eq:diffattn}
\small
\mathrm{DiffAttn}(\mathbf{Z})=(\mathbf{A}_1-\lambda\mathbf{A}_2)\mathbf{V},
\end{equation}
where $\lambda$ is a learnable scalar. The subtraction suppresses attention components that appear similarly in both maps and preserves token-level signals that are more distinctive. This matches the diagnosis in \cref{sec:2_bottleneck}: if standard attention becomes too diffuse under temporal shift, subtracting a common-mode attention component should increase contrast among historical tokens. Importantly, this intervention does not require changing the feature and sequence construction.

\subsection{Cross-architecture transferability}\label{sec:val_transfer}

We first test whether this attention-level change transfers across existing CTDG Transformers. We replace standard multi-head self-attention with differential attention in three representative baselines: \DyGFormer~\cite{yu2023towards}, \TIDFormer~\cite{Peng2025TIDFormer}, and \TCL~\cite{Wang2021TCL}. These models use different sequence construction strategies, including neighbor co-occurrence encoding, calendar-based temporal partitioning, and graph-topology-aware temporal encoding. For each model, all other components are kept unchanged, including the input sequence, feature channels, position or time encoding, and training procedure. We denote the resulting variants as \DyGFormer+DA, \TIDFormer+DA, and \TCL+DA.

\begin{table*}[!t]
\caption{\small Effect of adding Differential Attention (DA) to Transformer-based baselines the transductive setting. $\Delta$ denotes the absolute AP improvement after adding DA.}
\label{tab:da_plugin_transductive_ap}
\vspace{-1.5ex}
\centering
\scriptsize
\setlength{\tabcolsep}{2pt}
\resizebox{0.95\textwidth}{!}{%
\begin{tabularx}{\textwidth}{l*{9}{>{\centering\arraybackslash}X}c}
\toprule
Model & \Wikipedia & \Reddit & \UCI & \Enron & \Mooc & \CanParl & \USLegis & \UNTrade & \UNVote & Avg. $\Delta$ \\
\midrule
\TIDFormer 
& 99.30$\pm$0.03 & 99.38$\pm$0.03 & 97.44$\pm$0.10 & 93.17$\pm$0.06 & 93.42$\pm$0.15 & 99.84$\pm$0.09 & 66.51$\pm$2.56 & 60.84$\pm$1.09 & 69.01$\pm$1.24 & -- \\
\TIDFormer+DA 
& 99.37$\pm$0.03 & 99.46$\pm$0.02 & 98.00$\pm$0.11 & 95.18$\pm$0.11 & 95.37$\pm$0.63 & 99.87$\pm$0.31 & 76.23$\pm$0.62 & 84.13$\pm$1.02 & 82.36$\pm$0.54 & --  \\
$\Delta$ 
& +0.07 & +0.08 & +0.56 & +2.01 & +1.95 & +0.03 & +9.72 & +23.29 & +13.35 & +5.67 \\
\midrule
\DyGFormer 
& 98.81$\pm$0.02 & 98.82$\pm$0.06 & 95.51$\pm$0.20 & 91.33$\pm$0.16 & 86.48$\pm$0.06 & 97.36$\pm$0.45 & 71.11$\pm$0.59 & 66.46$\pm$1.29 & 55.55$\pm$0.42 & -- \\
\DyGFormer+DA 
& 98.89$\pm$0.09 & 99.17$\pm$0.05 & 96.18$\pm$0.01 & 94.12$\pm$0.15 & 95.21$\pm$0.26 & 97.41$\pm$0.03 & 82.15$\pm$0.34 & 90.73$\pm$0.17 & 81.94$\pm$0.77 & --  \\
$\Delta$ 
& +0.08 & +0.35 & +0.67 & +2.79 & +8.73 & +0.05 & +11.04 & +24.27 & +26.39 & +8.26 \\
\midrule
\TCL 
& 96.47$\pm$0.16 & 97.53$\pm$0.02 & 89.57$\pm$1.63 & 79.70$\pm$0.71 & 82.38$\pm$0.24 & 68.67$\pm$2.67 & 69.59$\pm$0.48 & 62.21$\pm$0.03 & 51.90$\pm$0.30 & -- \\
\TCL+DA 
& 97.52$\pm$0.37 & 98.32$\pm$0.08 & 91.53$\pm$0.54 & 85.72$\pm$0.23 & 85.41$\pm$0.40 & 71.28$\pm$0.95 & 73.26$\pm$0.69 & 66.04$\pm$0.12 & 54.80$\pm$0.08 & -- \\
$\Delta$ 
& +1.05 & +0.79 & +1.96 & +6.02 & +3.03 & +2.61 & +3.67 & +3.83 & +2.90 & +2.87 \\
\bottomrule
\end{tabularx}%
}
\vspace{-2ex}
\end{table*}

Tab.~\ref{tab:da_plugin_transductive_ap} reports the transductive AP results. Adding differential attention improves all three baselines, with average gains of $+8.26$ for \DyGFormer, $+5.67$ for \TIDFormer, and $+2.87$ for \TCL. The gains are concentrated on the high-shift datasets identified in \cref{sec:2_measure_gap}. For example, \DyGFormer+DA improves over \DyGFormer by $+11.04$, $+24.27$, and $+26.39$ AP on \USLegis, \UNTrade, and \UNVote, respectively. \TIDFormer+DA shows the same pattern, with gains of $+9.72$, $+23.29$, and $+13.35$ on the same datasets. In contrast, on low-shift datasets such as \Wikipedia and \Reddit, where the original baselines are already near saturation, the gains are small.

These results support the diagnosis in two ways. First, the improvement appears across architectures with different sequence construction designs, while the only modified component is attention. Second, the gains are largest exactly where temporal shift is strongest. This pattern would be difficult to explain if the main limitation were a model-specific input construction issue. It is more consistent with the hypothesis that attention allocation is a shared bottleneck under shift.

\subsection{Attention-level mechanism verification}\label{sec:val_mechanism}

We next test whether differential attention changes attention behavior in the way predicted by~\cref{sec:2_bottleneck}. Since differential attention forms a signed map $\mathbf{B}=\mathbf{A}_1-\lambda\mathbf{A}_2$, entropy and attention-mass measurements require a nonnegative normalized map. We therefore compute attention statistics using the normalized positive component {\small $\widehat{\mathbf{A}}^{\mathrm{DA}}_{ij} = \frac{[\mathbf{B}_{ij}]_{+}}{\sum_{\ell}[\mathbf{B}_{i\ell}]_{+}+\epsilon},$}
where $[\cdot]_+=\max(\cdot,0)$ and $\epsilon$ is a small constant for numerical stability. For standard attention, we use the usual softmax map. This gives a comparable probability distribution over historical tokens for both attention mechanisms.

\textbf{Attention entropy.} For each test query, we compute the Shannon entropy $H(\mathbf{a}) = -\sum_i a_i \log a_i$
over historical tokens, averaged across heads and test examples. Lower entropy indicates more concentrated attention. Tab.~\ref{tab:attention_focus} compares entropy at each layer when using differential vs.\ standard attention, with all other components identical. The pattern matches the prediction. Differential attention produces lower entropy than standard attention on most datasets, with especially large reductions on shifted datasets. For example, entropy decreases from $2.91$ to $2.31$ on \USLegis, from $3.23$ to $2.35$ on \UNVote, and from $3.91$ to $2.87$ on \UNTrade. On low-shift datasets where standard attention is already relatively concentrated, the gap is generally smaller. This supports the first mechanism prediction: differential attention reduces attention dispersion, and the reduction is strongest where the diagnosis predicts dispersion to be most harmful.

\begin{table*}[!t]
\caption{\small Comparison of attention entropy between standard attention and differential attention in the last layer. The reported values are averaged over attention heads.}
\label{tab:attention_focus}
\vspace{-1.5ex}
\centering
\scriptsize
\setlength{\tabcolsep}{4pt}
\resizebox{0.98\textwidth}{!}{%
\begin{tabularx}{\textwidth}{cc*{9}{>{\centering\arraybackslash}X}}
\toprule
Attention & Wikipedia & Reddit & UCI & Enron & Mooc & CanParl & USLegis & UNTrade & UNVote \\
\midrule
Differential Attention & 2.8075 & 2.9179 & 3.1967 & 3.0203 & 2.5780 & 2.7699 & 2.3099 & 2.8705 & 2.3461 \\
Standard Attention & 3.2894 & 3.8371 & 3.8586 & 3.3862 & 3.9390 & 3.6248 & 2.9130 & 3.9199 & 3.2337 \\
\bottomrule
\end{tabularx}%
}
\vspace{-2ex}
\end{table*}

\textbf{Attention mass on critical nodes.}
Lower entropy alone is not sufficient, since attention could become sharper on the wrong tokens. We therefore measure whether the recovered concentration falls on the critical nodes defined in~\cref{sec:2_critical_node}. We use two statistics. The first is the total attention mass assigned to critical nodes. The second is the proportion of critical nodes among the top-ranked attention tokens, with the top-$5\%$ region reflecting the model's most confident attention assignments.

\begin{table*}[!t]
\caption{\small Proportion of attention scores assigned to critical nodes in the last layer.}
\label{tab:critical_node_attention}
\vspace{-1.5ex}
\centering
\scriptsize
\setlength{\tabcolsep}{5pt}
\resizebox{0.98\textwidth}{!}{%
\begin{tabularx}{\textwidth}{l*{9}{>{\centering\arraybackslash}X}}
\toprule
Attention & Wikipedia & Reddit & UCI & Enron & Mooc & CanParl & USLegis & UNTrade & UNVote \\
\midrule
Differential Attention & \textbf{0.8006} & \textbf{0.7866} & \textbf{0.7951} & \textbf{0.9873} & \textbf{0.7958} & \textbf{0.8187} & \textbf{0.7800} & \textbf{0.9706} & \textbf{0.9801} \\
Standard Attention & 0.7692 & 0.7777 & 0.7315 & 0.9836 & 0.7840 & 0.7865 & 0.7768 & 0.7445 & 0.8255 \\
\bottomrule
\end{tabularx}%
}
\vspace{-1.5ex}
\end{table*}

\begin{table*}[!t]
\caption{\small Proportion of critical nodes among the top-5\%-ranked attention tokens in the last layer.}
\label{tab:critical_nodes_topk_attention}
\vspace{-1.5ex}
\centering
\scriptsize
\setlength{\tabcolsep}{4pt}
\resizebox{0.98\textwidth}{!}{%
\begin{tabularx}{\textwidth}{ll*{9}{>{\centering\arraybackslash}X}c}
\toprule
Attention & \Wikipedia & \Reddit & \UCI & \Enron & \Mooc & \CanParl & \USLegis & \UNTrade & \UNVote & Avg. \\
\midrule
 Differential Attention 
& \textbf{0.9395} & \textbf{0.9055} & \textbf{0.8990} & \textbf{0.9880} & \textbf{0.8985} & \textbf{0.8840} & \textbf{0.7845} & \textbf{0.9710} & \textbf{0.9810} & \textbf{0.9168} \\
Standard Attention 
& 0.7850 & 0.8945 & 0.6785 & 0.9825 & 0.8850 & 0.8220 & 0.7810 & 0.8470 & 0.8875 & 0.8403 \\
\bottomrule
\end{tabularx}%
}
\vspace{-2ex}
\end{table*}

Tab.~\ref{tab:critical_node_attention} shows that differential attention assigns more total mass to critical nodes than standard attention on all 9 datasets, with large gains on shifted datasets such as \UNTrade and \UNVote. Tab.~\ref{tab:critical_nodes_topk_attention} further shows that critical nodes are more frequent among the top-$5\%$ attended tokens under differential attention, increasing the average ratio from $0.84$ to $0.92$. Thus, differential attention does not merely sharpen attention arbitrarily. It shifts attention toward the same structurally and temporally important nodes whose predictive role was established by the ablation study in~\cref{sec:2_critical_node}.

Together, the plug-in experiments and attention-level measurements validate the diagnostic chain. Replacing only attention improves three different CTDG Transformers, with the largest gains on high-shift datasets. The same replacement also reduces attention entropy and increases focus on critical nodes. These results support the central conclusion of~\cref{sec:2}: the shared failure mode is not simply missing historical information or insufficient sequence construction, but dispersed attention allocation under temporal distribution shift.

\section{DiffDyG: A Reference Implementation}\label{sec:method}

The validation in~\cref{sec:validation} shows that replacing standard attention with differential attention improves several CTDG Transformers under temporal distribution shift. We now instantiate this finding in a single model, \DiffDyG, which serves as the configuration used for end-to-end comparison with state-of-the-art baselines in~\cref{sec:experiments}. \DiffDyG keeps the overall structure of a standard Transformer-based CTDG model, while replacing self-attention with differential attention and using standard temporal and structural encodings for dynamic graph inputs. Specifically, differential attention provides stronger temporal dispersion ability, allowing \DiffDyG to use a smaller embedding dimension than existing Transformer-based baselines. 

\textbf{Architecture overview. }
\DiffDyG consists of three components. First, it constructs a historical token sequence for each query node from its recent interactions. Second, each token combines node, edge, temporal, co-occurrence, and structural information. Third, the resulting sequence is processed by stacked Transformer layers whose self-attention modules are replaced with differential attention. Source and destination nodes are encoded separately, then combined for link prediction.

\subsection{Input construction}
\label{sec:input_construction}
We describe the input construction for the source node $u$ at time $t$. The destination node $v$ is processed in the same way.

\textbf{Token sequence.}
For source node $u$, we form a sequence of $K{+}1$ tokens. The first token represents $u$ itself, and the remaining $K$ tokens represent its most recent interaction neighbors before time $t$, ordered chronologically. Unless otherwise stated, we use $K=20$ first-hop neighbors. For three smaller benchmarks (\Wikipedia, \UCI, \CanParl) we additionally include 5 second-hop neighbors. Details and motivation are in Appendix~\ref{app:multihop}.

\textbf{Feature channels.} Each token aggregates five channels: 
(i) the \emph{node feature} of $u$ (or its neighbor $v'_j$ for $j > 0$); 
(ii) the \emph{edge feature} of the corresponding interaction; 
(iii) a \emph{temporal encoding} of the elapsed time $\Delta t_j = t - t_j$ via cosine basis with learnable frequencies; 
(iv) the \emph{co-occurrence frequency} of $v'_j$ in the histories of both $u$ and the destination $v$, following~\cite{yu2023towards}; 
(v) the \emph{spatial distance} (hop count) from $u$. Each channel is linearly projected to a common dimension $d$ and the five projections are concatenated into a $5d$-dimensional token representation. Stacking the tokens yields the input matrix $\mathbf{Z}_u^t \in \mathbb{R}^{(K+1) \times 5d}$. Specifically, for the source node $u$ (the 0-th token), its edge feature and temporal encoding feature are $\mathbf{0}$, and its spatial distance is $0$.

\textbf{Positional encoding.}
We apply Rotary Positional Embeddings (RoPE)~\cite{su2024roformer} to the token sequence before attention. RoPE provides relative position information within the historical sequence, while the elapsed-time encoding captures the actual temporal interval between the query time and each historical interaction.

\subsection{Differential attention encoder}
\label{sec:method_training}

Source and destination input matrices $\mathbf{Z}_u^t$ and $\mathbf{Z}_v^t$ are processed independently through $L = 2$ stacked differential attention transformer layers. Each layer contains a differential attention sublayer followed by a SwiGLU feed-forward network, with RMSNorm before each sublayer and residual connections around each. The output sequences are average-pooled along the token dimension to produce node-level embeddings $\mathbf{Y}_u^t, \mathbf{Y}_v^t \in \mathbb{R}^{d_{\mathrm{out}}}$, which are concatenated and passed through an MLP classifier to produce the link probability $p_{u,v}^t$. The model is trained end-to-end with binary cross-entropy loss using one negative edge per positive edge. The full training procedure is summarized in Algorithm~\ref{alg:diffdyg} in the appendix.
\begin{table*}[!t]
\caption{\small Performance (Average Precision) comparison in the transductive setting under random negative sampling strategies. Average Rank (AR) is computed across the 9 datasets. The best and second best results are shown in \textbf{bold} and  \underline{underlined}.}
\label{tab:ap_transductive_rand}
\vspace{-1.5ex}
\centering
\scriptsize
\setlength{\tabcolsep}{2pt}
\resizebox{0.98\textwidth}{!}{%
\begin{tabularx}{\textwidth}{l*{9}{>{\centering\arraybackslash}X}c}
\toprule
Baseline & \Wikipedia & \Reddit & \UCI & \Enron & \Mooc & \CanParl & \USLegis & \UNTrade & \UNVote & AR \\
\midrule
\DiffDyG & \textbf{99.57$\pm$0.03} & \textbf{99.64$\pm$0.14} & \textbf{98.54$\pm$0.03} & \textbf{98.77$\pm$0.46} & \textbf{97.01$\pm$0.22} & \underline{99.63$\pm$0.09} & \textbf{87.52$\pm$0.47} & \textbf{98.97$\pm$0.01} & \textbf{88.75$\pm$1.45} & \textbf{1.11} \\
\RepeatMixer & 99.16$\pm$0.02 & 99.22$\pm$0.01 & 96.74$\pm$0.08 & 92.66$\pm$0.07 & 92.76$\pm$0.10 & 81.30$\pm$0.76 & 68.36$\pm$0.28 & 64.00$\pm$0.06 & 55.33$\pm$0.07 & 4.89 \\
\TIDFormer & \underline{99.30$\pm$0.03} & \underline{99.38$\pm$0.03} & \underline{97.44$\pm$0.10} & \underline{93.17$\pm$0.06} & 93.42$\pm$0.15 & \textbf{99.84$\pm$0.09} & 66.51$\pm$2.56 & 60.84$\pm$1.09 & \underline{69.01$\pm$1.24} & 3.78 \\
\DyGFormer & 98.81$\pm$0.02 & 98.82$\pm$0.06 & 95.51$\pm$0.20 & 91.33$\pm$0.16 & 86.48$\pm$0.06 & 97.36$\pm$0.45 & 71.11$\pm$0.59 & 66.46$\pm$1.29 & 55.55$\pm$0.42 & 4.78 \\
\CNEN & 99.09$\pm$0.04 & 99.22$\pm$0.01 & 96.85$\pm$0.08 & 92.48$\pm$0.10 & \underline{94.18$\pm$0.07} & 86.09$\pm$0.14 & \underline{78.69$\pm$0.12} & \underline{76.92$\pm$0.06} & 66.40$\pm$0.12 & \underline{3.00} \\
\TGAT & 96.94$\pm$0.06 & 98.52$\pm$0.02 & 79.63$\pm$0.70 & 71.12$\pm$0.97 & 85.84$\pm$0.15 & 70.73$\pm$0.72 & 68.52$\pm$3.16 & 61.47$\pm$0.18 & 52.21$\pm$0.98 & 8.78 \\
\TCL & 96.47$\pm$0.16 & 97.53$\pm$0.02 & 89.57$\pm$1.63 & 79.70$\pm$0.71 & 82.38$\pm$0.24 & 68.67$\pm$2.67 & 69.59$\pm$0.48 & 62.21$\pm$0.03 & 51.90$\pm$0.30 & 9.22 \\
\TGN & 98.45$\pm$0.06 & 98.63$\pm$0.06 & 92.34$\pm$1.04 & 86.53$\pm$1.11 & 89.15$\pm$1.60 & 70.88$\pm$2.34 & 75.99$\pm$0.58 & 65.03$\pm$1.37 & 65.72$\pm$2.17 & 5.89 \\
\CAWN & 98.76$\pm$0.03 & 99.11$\pm$0.01 & 95.18$\pm$0.06 & 89.56$\pm$0.09 & 80.15$\pm$0.25 & 69.82$\pm$2.34 & 70.58$\pm$0.48 & 65.39$\pm$0.12 & 52.84$\pm$0.10 & 6.67 \\
\EdgeBank & 90.37$\pm$0.00 & 94.86$\pm$0.00 & 76.20$\pm$0.00 & 83.53$\pm$0.00 & 57.97$\pm$0.00 & 64.55$\pm$0.00 & 58.39$\pm$0.00 & 60.41$\pm$0.00 & 58.49$\pm$0.00 & 10.00 \\
\GraphMixer & 97.25$\pm$0.03 & 97.31$\pm$0.01 & 93.25$\pm$0.57 & 82.25$\pm$0.16 & 82.78$\pm$0.15 & 77.04$\pm$0.46 & 70.74$\pm$1.02 & 62.61$\pm$0.27 & 52.11$\pm$0.16 & 7.78 \\
\bottomrule
\end{tabularx}%
}
\vspace{-2ex}
\end{table*}

\section{Experimental Evaluation of \DiffDyG}\label{sec:experiments}

We evaluate whether \DiffDyG translates the attention-level fix validated in~\cref{sec:validation} into end-to-end performance gains. The main text focuses on transductive link prediction under random negative sampling, which is the standard setting used for direct comparison. Full results for AUC-ROC, inductive evaluation, historical negative sampling, inductive negative sampling, efficiency, and additional analyses and extension to dynamic node prediction task are reported in Appendix~\ref{app:additional_exp}.

\paragraph{Setup.} We use the 9 CTDG benchmarks introduced in~\cref{sec:2}, with chronological 70\%/15\%/15\% train/validation/test splits. We report Average Precision (AP) averaged over 5 random seeds. Baselines include memory-based models (\TGN~\cite{rossi2020temporal}, \EdgeBank~\cite{poursafaei2022towards}), random-walk-based models (\CAWN~\cite{wang2021inductive}), temporal message-passing transformers (\TGAT~\cite{xu2020inductive}, \TCL~\cite{Wang2021TCL}, \DyGFormer~\cite{yu2023towards}, \TIDFormer~\cite{Peng2025TIDFormer}), sequence-based models (\GraphMixer~\cite{cong2023we}, \RepeatMixer~\cite{zou2024repeat}), and structure-encoding models (\CNEN~\cite{cheng2024co}). Full baseline descriptions and implementation details are in Appendix~\ref{app:baselines} and Appendix~\ref{app:implement}.

\subsection{Performance comparison with baselines}\label{sec:exp_main}

Tab.~\ref{tab:ap_transductive_rand} reports the transductive AP results under random negative sampling. \DiffDyG achieves the best average rank of $1.11$, ranking first on 7 out of 9 datasets and second on the remaining 2. The strongest gains appear on the high-shift datasets identified in~\cref{sec:2_measure_gap}. On \USLegis, \DiffDyG outperforms the second-best baseline from $78.69\%$ to $87.52\%$. On \UNTrade, it improves from $76.92\%$ to $98.97\%$. On \UNVote, it improves from $69.01\%$ to $88.75\%$. These are precisely the datasets where existing CTDG Transformers plateaued despite changes to sequence construction and feature design.

On low-shift datasets such as \Wikipedia and \Reddit, where existing methods are already near saturation, \DiffDyG preserves performance, reaching $99.57\%$ and $99.64\%$ AP. On medium-shift datasets such as \UCI and \Enron, it further improves AP to $98.54\%$ and $98.77\%$. Thus, the attention-level fix does not trade off performance on easier benchmarks for gains on shifted ones.

The gain pattern matches the diagnosis in~\cref{sec:2}. Improvements are small on saturated low-shift datasets, moderate on medium-shift datasets, and largest on high-shift datasets. This is consistent with the view that attention dispersion becomes most harmful under temporal distribution shift, and that differential attention primarily helps by restoring sharper allocation over informative historical tokens. 

The full results in Appendix~\ref{app:additional_exp:sampling} show the same trend. DiffDyG achieves the best average ranking across all three negative sampling strategies for both AP and AUC-ROC in both transductive and inductive settings, confirming that its advantage is not tied to a specific evaluation protocol.

\begin{table*}
\caption{\small Ablation study on 9 datasets in transductive settings. Results are reported in Average Precision (AP)}
\label{table:acc:ablation}
\vspace{-4pt}
\centering
\setlength{\tabcolsep}{3pt}
\scriptsize
\resizebox{0.95\textwidth}{!}{
\begin{tabularx}{\textwidth}{l*{9}{>{\centering\arraybackslash}X}}
\toprule
AP & \Wikipedia & \Reddit & \UCI & \Enron & \Mooc & \CanParl & \USLegis & \UNTrade & \UNVote \\
\midrule
\DiffDyG & 99.57$\pm$0.03 & 99.64$\pm$0.14 & 98.54$\pm$0.03 & 98.77$\pm$0.46 & 97.01$\pm$0.22 & 99.63$\pm$0.09 & 87.52$\pm$0.47 & 98.97$\pm$0.01 & 88.75$\pm$1.45 \\
w/o RoPE & 99.02$\pm$0.04 & 99.16$\pm$0.02 & 97.14$\pm$2.00 & 97.81$\pm$3.43 & 96.37$\pm$2.09 & 97.56$\pm$0.30 & 85.09$\pm$2.41 & 95.50$\pm$4.00 & 84.98$\pm$2.72 \\
w/o DA & 98.98$\pm$0.36 & 99.02$\pm$0.15 & 96.94$\pm$0.29 & 92.83$\pm$0.24 & 88.61$\pm$0.38 & 97.42$\pm$0.17 & 74.51$\pm$0.08 & 68.11$\pm$0.32 & 57.21$\pm$0.27 \\
w/o SE & 99.27$\pm$0.35 & 99.34$\pm$0.32 & 98.06$\pm$0.01 & 98.15$\pm$0.65 & 95.87$\pm$0.06 & 98.87$\pm$0.01 & 85.89$\pm$0.37 & 96.23$\pm$0.58 & 86.43$\pm$0.01 \\
\bottomrule
\end{tabularx}
}
\vspace{-2ex}
\end{table*}

\subsection{Component ablation}
\label{sec:exp_ablation}

We next examine which components drive \DiffDyG's gains. We compare the full model against three variants: \emph{w/o DA}, which replaces differential attention with standard multi-head attention; \emph{w/o RoPE}, which removes rotary positional embeddings; and \emph{w/o SE}, which removes spatial distance encoding. The full dataset-wise ablation is reported in Appendix~\ref{app:ablation}; the main text summarizes the key effects.

Removing differential attention causes the largest drop, with an average AP decrease of $10.53$ points across the 9 datasets. The drop is especially severe on high-shift datasets: $-13.01$ on \USLegis, $-30.86$ on \UNTrade, and $-31.54$ on \UNVote. In contrast, removing RoPE causes an average drop of $1.75$ points, and removing spatial encoding causes an average drop of $1.14$ points. This confirms within \DiffDyG what~\cref{sec:val_transfer} showed across architectures: differential attention is the dominant source of improvement, while the supporting encodings provide smaller refinements.

\section{Discussion and Limitations}
\label{sec:conclusion}

This work studied why Transformer-based CTDG models underperform on benchmarks with large temporal distribution shift. Instead of focusing on sequence construction or feature design, we examined whether the bottleneck lies in the attention mechanism itself. Our diagnosis shows that shifted datasets still contain predictive historical signals, captured by structurally and temporally important critical nodes, but standard attention does not allocate sufficient mass to these nodes. The failure is therefore not simply missing information or insufficient model capacity. It is an attention-allocation problem: under shift, attention becomes too dispersed to reliably separate predictive historical neighbors from less informative ones.

We validated this diagnosis by using differential attention as a targeted intervention. Replacing only the attention module in three existing CTDG Transformers consistently improved performance, with the largest gains on the high-shift datasets identified by the diagnosis. Direct attention-level measurements further showed that differential attention reduces entropy and increases attention mass on critical nodes. Building on these findings, \DiffDyG provides a reference implementation of this attention-level fix and achieves SOTA results across the standard CTDG benchmarks. These results suggest that further progress on difficult CTDG datasets may require not only better historical sequence construction, but also better mechanisms for allocating attention over those sequences.

\textbf{Limitations.}
Our analysis is primarily empirical. Although the ablation studies, plug-in experiments, and attention measurements provide consistent evidence for attention dispersion, we do not provide a formal characterization of when softmax attention becomes dispersed under temporal shift or when differential attention is guaranteed to help. Developing such theory is an important direction for future work. In addition, our definition of critical nodes is a diagnostic probe based on structural centrality and temporal stability, rather than a unique or optimal definition of predictive historical neighbors. Other definitions, including learned or task-adaptive ones, may reveal additional signals. Finally, our evaluation focuses on dynamic link prediction on standard CTDG benchmarks. Extending the diagnosis to other dynamic graph tasks and to substantially larger industrial graphs remains future work.

\bibliographystyle{abbrvnat}
\bibliography{ref}


\appendix


\newpage
\section{Additional experimental details}\label{app:exp_details}

\subsection{Details of datasets}\label{app:datasets}

\begin{table}[H]
\centering
\scriptsize
\caption{\small Summary of dynamic graph datasets}\label{Tab:dataset_details}
\resizebox{0.99\textwidth}{!}{
\begin{tabular}{l c r r c c r c r}
\toprule
\textbf{Datasets} & \textbf{Domains} & \textbf{\#Nodes} & \textbf{\#Links} & \textbf{\#Node \& Link Feat.} & \textbf{Bipartite} & \textbf{Duration} & \textbf{Time Granularity} & \textbf{\# Steps} \\
\midrule
\Wikipedia     & Social      & 9,227  & 157,474   & -- \& 172 & \cmark  & 1 month     & Unix timestamps & 152,757    \\
\Reddit        & Social      & 10,984 & 672,447   & -- \& 172 & \cmark  & 1 month     & Unix timestamps & 669,065    \\
\Mooc          & Interaction & 7,144  & 411,749   & -- \& 4   & \cmark  & 17 months   & Unix timestamps & 345,600    \\
\Enron         & Social      & 184    & 125,235   & -- \& --  & \xmark & 3 years     & Unix timestamps & 22,632     \\
\UCI           & Social      & 1,899  & 59,835    & -- \& --  & \xmark & 196 days    & Unix timestamps & 58,911     \\
\CanParl       & Politics    & 734    & 74,478    & -- \& 1   & \xmark & 14 years    & years            & 14         \\
\USLegis       & Politics    & 225    & 60,396    & -- \& 1   & \xmark & 12 congresses & congresses     & 12         \\
\UNTrade       & Economics   & 255    & 507,497   & -- \& 1   & \xmark & 32 years    & years            & 32         \\
\UNVote        & Politics    & 201    & 1,035,742 & -- \& 1   & \xmark & 72 years    & years            & 72         \\
\bottomrule
\end{tabular}}
\end{table}

We conduct experiments on 9 widely used dynamic graph benchmarks. These datasets are collected~\citet{poursafaei2022towards}. Tab.~\ref{Tab:dataset_details} provides a summary of their key statistics. Below we describe each dataset.

\textbf{\Wikipedia} is a bipartite dataset that captures editing activities on Wikipedia during one month. The nodes represent users and pages. Each edge corresponds to an edit action at a specific timestamp and is associated with a 172-dimensional LIWC feature. The dataset also includes dynamic labels that indicate whether a user was temporarily banned from editing.

\textbf{\Reddit} is another bipartite graph from the social domain. It captures posting activity of users on different subreddits during one month. The nodes represent users and subreddits. Each edge denotes a posting event at a specific timestamp and is associated with a 172-dimensional LIWC feature vector. Dynamic labels indicate whether a user was banned from posting.

\textbf{\Mooc} is a bipartite graph constructed from interactions on an online education platform. The nodes represent students and course units such as videos or problem sets. Each edge represents a student accessing a course unit at a specific timestamp and includes a 4-dimensional feature vector.

\textbf{\Enron} is a non-bipartite graph that captures email communication among Enron employees over three years. The nodes represent employees. Each edge corresponds to an email exchange at a recorded timestamp.

\textbf{\UCI} is a non-bipartite graph that records message exchanges among university students over approximately 196 days. The nodes represent students. Each edge represents a directed message sent at a specific timestamp.

\textbf{\CanParl} is a political graph that tracks voting behavior among Canadian Members of Parliament (MPs) from 2006 to 2019. The nodes represent Members of Parliament. An edge exists between two MPs in a given year when both vote “yes” on the same bill. The edge weight equals the number of such shared “yes” votes in that year.

\textbf{\USLegis} is a co-sponsorship network of the U.S. Senate, spanning eight years. The nodes represent senators. An edge exists when two legislators co-sponsor the same bill. The weight of an edge reflects the number of cosponsorships within one congressional session.

\textbf{\UNTrade} is a long-term trade network that captures agricultural and food exchanges among 181 countries over 32 years. The edge weight between two countries reflects the total normalized value of imports and exports between them in this domain.

\textbf{\UNVote} is a voting network of the United Nations General Assembly over 72 years. For each resolution, an edge weight between two countries increases by one when both cast a “yes” vote on that resolution.

\subsection{Details of Baselines}
\label{app:baselines}
\textbf{\TGAT}~\cite{xu2020inductive} learns node embeddings by attending over each node’s temporal–topological neighbors via a self-attention framework, coupled with a time-encoding function that captures temporal patterns in dynamic graphs.

\textbf{\TCL}~\cite{Wang2021TCL} processes dynamic graphs by first utilizing a graph-topology-aware transformer to capture temporal and topological information. It then employs a two-stream encoder to separately extract representations from the temporal neighborhoods of two interacting nodes. A co-attentional transformer models the inter-dependencies between these nodes at a semantic level, enhanced by contrastive learning.

\textbf{\TGN}~\cite{rossi2020temporal} maintains an evolving memory for every node and updates it whenever a new interaction arrives, using a message function, a message aggregator, and a memory updater; an embedding module then produces time-aware node representations.

\textbf{\CAWN}~\cite{wang2021inductive} first samples multiple causal anonymous walks per node to capture the dynamics and relative identities, encodes each walk with recurrent networks, and aggregates the walk encodings to form the final node representation for downstream tasks.

\textbf{\EdgeBank}~\cite{poursafaei2022towards} is a parameter-free, memory-based method tailored to transductive dynamic link prediction. It predicts an interaction as positive if it has been retained in memory and negative otherwise, operating purely over stored historical edges. 

\textbf{\GraphMixer}~\cite{cong2023we} shows that a fixed (non-trainable) time encoding can outperform a learned one. It embeds temporal links with an MLP-Mixer–style link encoder and summarizes node features via neighbor mean-pooling.

\textbf{\DyGFormer}~\cite{yu2023towards} is Transformer-based and emphasizes source–target correlation modeling through a neighbor co-occurrence encoding over historical sequences. It further introduces a patching trick that splits long histories into patches for efficient, effective learning. 

\textbf{\RepeatMixer}~\cite{zou2024repeat} learns temporal interaction patterns with an MLP encoder driven by an evolving repeat-behavior sampling scheme (covering first- and higher-order repeats), and adopts a time-aware aggregation that adaptively weighs representations from different orders according to their temporal salience.

\textbf{CNEN}~\cite{cheng2024co} proposes a lightweight dynamic graph model that stores structural signals in a hashtable-based co-neighbor memory with short- and long-term components; it fuses node, edge, temporal, and structural cues via MLPs to produce time-aware node embeddings for efficient link prediction.

\textbf{\TIDFormer}~\cite{Peng2025TIDFormer} is a Transformer architecture that leverages calendar-based time partitioning and derives informative interaction embeddings using only sampled first-order neighbors on both bipartite and non-bipartite graphs; a simple decomposition module tracks shifts in historical interaction patterns to model temporal and interactive factors.

\subsection{Details of Model Configurations, Implementation, and Evaluation}\label{app:implement}

For \DiffDyG, we set the number of Transformer layers to 2, the number of attention heads to 2, and the dropout rate to 0.2. We set the dimension $d$ to 36 and the per-head attention dimension $d_{\text{attn}}$ to 45. By default, we sample 20 first-order neighbors for both the source and destination nodes to construct the input sequence. The hyperparameters of all baseline methods follow the default settings reported in their original papers.

To ensure a fair comparison, all models are trained for up to 100 epochs with early stopping based on a patience of 5 epochs. We use the Adam optimizer with a learning rate of $1 \times 10^{-4}$, a batch size of 200, and a weight decay of $1 \times 10^{-4}$. All experiments are conducted on a single NVIDIA GeForce RTX 4090 GPU. All experiments use 5 random seeds and we report mean and standard deviation.

The default hyperparameter settings of our method are summarized in Tab.~\ref{tab:hp_ranges}. For negative sampling, we adopt random negative sampling by default.


\subsubsection{Multi-Hop Neighborhood Extension}\label{app:multihop}

The input construction described in~\cref{sec:input_construction} uses only the direct (1-hop) interaction neighbors of each node. In some smaller-scale datasets (namely, Wikipedia, UCI, and Can.\ Parl.), the 1-hop interaction is relatively sparse. Therefore, we apply a straightforward extension to $2$-hop neighborhoods for these 3 datasets, where the additional computational cost is manageable.

\textbf{Extended token sequence.}  
Starting from the 1-hop neighbors $\{(v'_1, t_1), \ldots, (v_K, t_K)\}$ of source node $u$, we recursively expand: for each hop level $h = 2, \ldots, H$, we sample $K_h$ most recent interaction neighbors of each $(h{-}1)$-hop neighbor, excluding nodes already in the sequence. The full token sequence then becomes
\begin{equation}
\mathbf{Z}^t_u = 
\bigl[\,\underbrace{\mathbf{z}^{(0)}}_{\text{source}}\;;\;
\underbrace{\mathbf{z}^{(1)},\ldots,\mathbf{z}^{(K)}}_{\text{1-hop}}\;;\;
\underbrace{\mathbf{z}^{(K+1)},\ldots}_{\text{2-hop}}\;;\;
\cdots\,\bigr]
\;\in\; \mathbb{R}^{(1 + K + K_2 + \cdots + K_H) \times 5d}.
\end{equation}

\textbf{Per-token features.}  
The five feature channels are computed identically to the 1-hop case,  with one natural change: the spatial distance embedding  $\mathbf{k}^{(j)}_S$ now encodes hop distance $h \in \{0, 1, \ldots, H\}$ rather than being uniformly 1 for all neighbors. This allows the attention mechanism to distinguish neighbors at different structural distances from the source node.

\textbf{Ordering.}  
Within each hop level, tokens are ordered chronologically by interaction time. Hop levels are concatenated in increasing order (1-hop before 2-hop, etc.), so the sequence reflects both temporal recency and structural proximity.

\textbf{Computational considerations.}  
Obtaining multi-hop samples can be expensive for large and dense graphs. Therefore, we only apply 2-hop sampling on the three datasets. Additionally, multi-hop sampling increases the sequence length and, consequently, the cost of attention, which scales quadratically. 

In our experiments, we sample $K{=}20$ first-hop and $K_2{=}5$ second-hop neighbors ($H{=}2$), extending the sequence from 21 to 26 tokens per node. This overhead is modest. For the remaining six larger datasets, we use 1-hop sampling only, and the performance difference is minor (see Tab.~\ref{tab:hyperparameter_trans} and \ref{tab:hyperparameter_ind} for the hop ablation).

\begin{table}[ht]
\centering
\caption{\small Default Hyperparameters.}
\footnotesize
\label{tab:hp_ranges}
\begin{tabular}{l l}
\toprule
\textbf{Hyperparameters} & \textbf{Values} \\
\midrule
Number of Transformer Layers $L$ &
$2$ \\
\addlinespace[2pt]
Number of Attention Heads $H$&
$2$  \\
\addlinespace[2pt]
Dimension of Time Intervals Encoding $d_T$ &
$100$ \\
\addlinespace[2pt]
Dimension of Interaction Frequency $d_C$ &
$36$ \\
\addlinespace[2pt]
Dimension of Structural Distance Encoding $d_S$ &
$1$ \\
\addlinespace[2pt]
$d$ & 
$36$ \\
\addlinespace[2pt]
$d_{attn}$ & 
$45$ \\
\addlinespace[2pt]
Number of Sampled 1-hop Neighbors $K$ &
$20$ \\
\addlinespace[2pt]
Dropout Rate &
$0.2$ \\
\addlinespace[2pt]
Learning Rate &
$0.0001$ \\
\addlinespace[2pt]
Batch Size &
$200$ \\
\addlinespace[2pt]
Weight Decay &
$0.0001$ \\
\addlinespace[2pt]
\bottomrule
\end{tabular}
\end{table}

\begin{table*}[!t]
\caption{\small Performance (Average Precision) comparison in the transductive setting. Average Rank (AR) is computed across the 9 datasets. The best and second best results are shown in \textbf{bold} and  \underline{underlined}.}
\label{tab:ap_transductive_nss}
\centering
\scriptsize
\setlength{\tabcolsep}{2pt}
\textbf{Random Negative Sampling}\\[0.2em]
\resizebox{0.98\textwidth}{!}{%
\begin{tabularx}{\textwidth}{l*{9}{>{\centering\arraybackslash}X}c}
\toprule
Baseline & \Wikipedia & \Reddit & \UCI & \Enron & \Mooc & \CanParl & \USLegis & \UNTrade & \UNVote & AR \\
\midrule
\DiffDyG & \textbf{99.57$\pm$0.03} & \textbf{99.64$\pm$0.14} & \textbf{98.54$\pm$0.03} & \textbf{98.77$\pm$0.46} & \textbf{97.01$\pm$0.22} & \underline{99.63$\pm$0.09} & \textbf{87.52$\pm$0.47} & \textbf{98.97$\pm$0.01} & \textbf{88.75$\pm$1.45} & \textbf{1.11} \\
\RepeatMixer & 99.16$\pm$0.02 & 99.22$\pm$0.01 & 96.74$\pm$0.08 & 92.66$\pm$0.07 & 92.76$\pm$0.10 & 81.30$\pm$0.76 & 68.36$\pm$0.28 & 64.00$\pm$0.06 & 55.33$\pm$0.07 & 4.89 \\
\TIDFormer & \underline{99.30$\pm$0.03} & \underline{99.38$\pm$0.03} & \underline{97.44$\pm$0.10} & \underline{93.17$\pm$0.06} & 93.42$\pm$0.15 & \textbf{99.84$\pm$0.09} & 66.51$\pm$2.56 & 60.84$\pm$1.09 & \underline{69.01$\pm$1.24} & 3.78 \\
\DyGFormer & 98.81$\pm$0.02 & 98.82$\pm$0.06 & 95.51$\pm$0.20 & 91.33$\pm$0.16 & 86.48$\pm$0.06 & 97.36$\pm$0.45 & 71.11$\pm$0.59 & 66.46$\pm$1.29 & 55.55$\pm$0.42 & 4.78 \\
\CNEN & 99.09$\pm$0.04 & 99.22$\pm$0.01 & 96.85$\pm$0.08 & 92.48$\pm$0.10 & \underline{94.18$\pm$0.07} & 86.09$\pm$0.14 & \underline{78.69$\pm$0.12} & \underline{76.92$\pm$0.06} & 66.40$\pm$0.12 & \underline{3.00} \\
\TGAT & 96.94$\pm$0.06 & 98.52$\pm$0.02 & 79.63$\pm$0.70 & 71.12$\pm$0.97 & 85.84$\pm$0.15 & 70.73$\pm$0.72 & 68.52$\pm$3.16 & 61.47$\pm$0.18 & 52.21$\pm$0.98 & 8.78 \\
\TCL & 96.47$\pm$0.16 & 97.53$\pm$0.02 & 89.57$\pm$1.63 & 79.70$\pm$0.71 & 82.38$\pm$0.24 & 68.67$\pm$2.67 & 69.59$\pm$0.48 & 62.21$\pm$0.03 & 51.90$\pm$0.30 & 9.22 \\
\TGN & 98.45$\pm$0.06 & 98.63$\pm$0.06 & 92.34$\pm$1.04 & 86.53$\pm$1.11 & 89.15$\pm$1.60 & 70.88$\pm$2.34 & 75.99$\pm$0.58 & 65.03$\pm$1.37 & 65.72$\pm$2.17 & 5.89 \\
\CAWN & 98.76$\pm$0.03 & 99.11$\pm$0.01 & 95.18$\pm$0.06 & 89.56$\pm$0.09 & 80.15$\pm$0.25 & 69.82$\pm$2.34 & 70.58$\pm$0.48 & 65.39$\pm$0.12 & 52.84$\pm$0.10 & 6.67 \\
\EdgeBank & 90.37$\pm$0.00 & 94.86$\pm$0.00 & 76.20$\pm$0.00 & 83.53$\pm$0.00 & 57.97$\pm$0.00 & 64.55$\pm$0.00 & 58.39$\pm$0.00 & 60.41$\pm$0.00 & 58.49$\pm$0.00 & 10.00 \\
\GraphMixer & 97.25$\pm$0.03 & 97.31$\pm$0.01 & 93.25$\pm$0.57 & 82.25$\pm$0.16 & 82.78$\pm$0.15 & 77.04$\pm$0.46 & 70.74$\pm$1.02 & 62.61$\pm$0.27 & 52.11$\pm$0.16 & 7.78 \\
\bottomrule
\end{tabularx}%
}
\textbf{Historical Negative Sampling}\\[0.2em]
\resizebox{0.98\textwidth}{!}{%
\begin{tabularx}{\textwidth}{l*{9}{>{\centering\arraybackslash}X}c}
\toprule
Baseline & \Wikipedia & \Reddit & \UCI & \Enron & \Mooc & \CanParl & \USLegis & \UNTrade & \UNVote & AR \\
\midrule
\DiffDyG & 90.13$\pm$0.14 & \textbf{84.01$\pm$1.14} & \underline{89.21$\pm$0.16} & \underline{86.27$\pm$0.69} & \underline{92.80$\pm$0.49} & \textbf{97.15$\pm$0.37} & \textbf{86.80$\pm$0.03} & \underline{78.50$\pm$4.13} & \underline{82.69$\pm$0.99} & \textbf{1.89} \\
\RepeatMixer & 90.20$\pm$1.04 & 83.02$\pm$1.20 & 87.23$\pm$0.23 & \textbf{87.38$\pm$0.18} & 92.19$\pm$0.58 & 75.65$\pm$1.99 & 66.35$\pm$5.64 & 58.57$\pm$2.45 & 55.32$\pm$0.02 & 4.33 \\
\TIDFormer & \textbf{91.21$\pm$0.78} & \underline{83.27$\pm$1.01} & \textbf{89.57$\pm$1.17} & 81.59$\pm$1.17 & \textbf{95.37$\pm$1.09} & 66.35$\pm$9.00 & 69.65$\pm$0.30 & 60.06$\pm$1.17 & 60.64$\pm$0.90 & \underline{3.89} \\
\DyGFormer & 82.23$\pm$2.54 & 81.57$\pm$0.67 & 82.17$\pm$0.82 & 75.63$\pm$0.73 & 85.85$\pm$0.66 & \underline{97.00$\pm$0.31} & \underline{85.30$\pm$3.88} & 64.41$\pm$1.40 & 60.84$\pm$1.58 & 4.89 \\
\CNEN & 78.19$\pm$2.28 & 82.76$\pm$0.83 & 82.10$\pm$0.95 & 77.66$\pm$0.15 & 87.68$\pm$0.13 & 76.74$\pm$1.11 & 77.07$\pm$7.77 & 74.78$\pm$0.21 & 67.55$\pm$0.86 & 4.78 \\
\TGAT & 87.38$\pm$0.22 & 79.55$\pm$0.20 & 68.27$\pm$1.37 & 64.07$\pm$1.05 & 82.19$\pm$0.62 & 67.13$\pm$0.84 & 62.14$\pm$6.60 & 55.74$\pm$0.91 & 52.96$\pm$2.14 & 8.56 \\
\TCL & 89.05$\pm$0.39 & 77.14$\pm$0.16 & 80.25$\pm$2.74 & 70.66$\pm$0.39 & 77.06$\pm$0.41 & 65.93$\pm$3.00 & 80.53$\pm$3.95 & 55.90$\pm$1.17 & 52.30$\pm$2.35 & 8.11 \\
\TGN & 86.86$\pm$0.33 & 81.22$\pm$0.61 & 80.43$\pm$2.12 & 73.91$\pm$1.76 & 87.06$\pm$1.93 & 68.42$\pm$3.07 & 74.00$\pm$7.57 & 58.44$\pm$5.51 & 69.37$\pm$3.93 & 6.11 \\
\CAWN & 71.21$\pm$1.67 & 80.82$\pm$0.45 & 65.30$\pm$0.43 & 64.73$\pm$0.36 & 74.05$\pm$0.95 & 66.53$\pm$2.77 & 68.82$\pm$8.23 & 55.71$\pm$0.38 & 51.26$\pm$0.04 & 9.56 \\
\EdgeBank & 73.35$\pm$0.00 & 73.59$\pm$0.00 & 65.50$\pm$0.00 & 76.53$\pm$0.00 & 60.71$\pm$0.00 & 63.84$\pm$0.00 & 63.22$\pm$0.00 & \textbf{81.32$\pm$0.00} & \textbf{84.89$\pm$0.00} & 7.89 \\
\GraphMixer & \underline{90.90$\pm$0.10} & 78.44$\pm$0.18 & 84.11$\pm$1.35 & 77.98$\pm$0.92 & 77.77$\pm$0.92 & 74.34$\pm$0.87 & 81.65$\pm$1.02 & 57.05$\pm$1.22 & 51.20$\pm$1.60 & 6.00 \\
\bottomrule
\end{tabularx}%
}
\textbf{Inductive Negative Sampling}\\[0.2em]
\resizebox{0.98\textwidth}{!}{%
\begin{tabularx}{\textwidth}{l*{9}{>{\centering\arraybackslash}X}c}
\toprule
Baseline & \Wikipedia & \Reddit & \UCI & \Enron & \Mooc & \CanParl & \USLegis & \UNTrade & \UNVote & AR \\
\midrule
\DiffDyG & \underline{88.89$\pm$2.13} & \textbf{91.87$\pm$0.02} & \textbf{86.07$\pm$0.01} & \textbf{84.15$\pm$0.03} & \textbf{85.27$\pm$1.75} & \textbf{96.89$\pm$0.04} & \textbf{86.64$\pm$0.65} & \textbf{76.47$\pm$0.57} & \textbf{81.93$\pm$0.08} & \textbf{1.11} \\
\RepeatMixer & 88.86$\pm$0.97 & 91.11$\pm$0.73 & 84.20$\pm$0.34 & \underline{83.17$\pm$0.50} & 83.11$\pm$1.28 & 76.16$\pm$2.12 & 69.01$\pm$0.60 & 58.57$\pm$2.45 & 55.32$\pm$0.02 & 4.56 \\
\TIDFormer & \textbf{91.15$\pm$1.17} & 91.57$\pm$0.50 & \underline{85.81$\pm$1.39} & 79.67$\pm$0.66 & \underline{84.53$\pm$1.11} & 64.00$\pm$0.77 & 69.87$\pm$1.72 & 60.32$\pm$2.13 & 59.87$\pm$0.19 & \underline{4.44} \\
\DyGFormer & 78.29$\pm$5.38 & 91.11$\pm$0.40 & 72.25$\pm$1.71 & 77.41$\pm$0.89 & 81.24$\pm$0.69 & \underline{95.44$\pm$0.57} & \underline{81.25$\pm$3.62} & 55.79$\pm$1.02 & 51.91$\pm$0.84 & 5.67 \\
\CNEN & 74.99$\pm$3.07 & 90.76$\pm$0.29 & 69.19$\pm$1.61 & 74.88$\pm$0.23 & 80.42$\pm$0.23 & 75.87$\pm$1.23 & 72.27$\pm$8.11 & 72.28$\pm$0.46 & 66.09$\pm$0.68 & 5.78 \\
\TGAT & 87.00$\pm$0.16 & 89.59$\pm$0.24 & 68.67$\pm$0.84 & 63.94$\pm$1.36 & 75.95$\pm$0.64 & 68.82$\pm$1.21 & 61.91$\pm$5.82 & 60.61$\pm$1.24 & 52.89$\pm$1.61 & 7.78 \\
\TCL & 86.76$\pm$0.72 & 87.45$\pm$0.29 & 76.01$\pm$1.11 & 71.29$\pm$0.32 & 74.65$\pm$0.54 & 65.85$\pm$1.75 & 78.15$\pm$3.34 & 61.06$\pm$1.74 & 50.62$\pm$0.82 & 7.22 \\
\TGN & 85.62$\pm$0.44 & 88.10$\pm$0.24 & 70.94$\pm$0.71 & 70.89$\pm$2.72 & 77.50$\pm$2.91 & 65.34$\pm$2.87 & 67.57$\pm$6.47 & 61.04$\pm$6.01 & \underline{67.63$\pm$2.67} & 7.00 \\
\CAWN & 74.06$\pm$2.62 & \underline{91.67$\pm$0.24} & 64.61$\pm$0.48 & 75.15$\pm$0.58 & 73.51$\pm$0.94 & 67.75$\pm$1.00 & 65.81$\pm$8.52 & 62.54$\pm$0.67 & 52.19$\pm$0.34 & 7.33 \\
\EdgeBank & 80.63$\pm$0.00 & 85.48$\pm$0.00 & 57.43$\pm$0.00 & 73.89$\pm$0.00 & 49.43$\pm$0.00 & 62.16$\pm$0.00 & 64.74$\pm$0.00 & \underline{72.97$\pm$0.00} & 66.30$\pm$0.00 & 8.22 \\
\GraphMixer & 88.59$\pm$0.17 & 85.26$\pm$0.11 & 80.10$\pm$0.51 & 75.01$\pm$0.79 & 74.27$\pm$0.92 & 69.48$\pm$0.63 & 79.63$\pm$0.84 & 60.15$\pm$1.29 & 51.60$\pm$0.73 & 6.78 \\
\bottomrule
\end{tabularx}%
}
\end{table*}
\begin{table*}[!t]
\caption{\small Performance (AUC-ROC) comparison in the transductive setting. Average Rank (AR) is computed across the 9 datasets. The best and second best results are shown in \textbf{bold} and  \underline{underlined}.}
\label{tab:auc_transductive_nss}
\centering
\scriptsize
\setlength{\tabcolsep}{2pt}
\textbf{Random  Negative Sampling}\\[0.2em]
\resizebox{0.98\textwidth}{!}{%
\begin{tabularx}{\textwidth}{l*{9}{>{\centering\arraybackslash}X}c}
\toprule
Baseline & \Wikipedia & \Reddit & \UCI & \Enron & \Mooc & \CanParl & \USLegis & \UNTrade & \UNVote & AR \\
\midrule
\DiffDyG & \textbf{99.37$\pm$0.38} & 99.07$\pm$1.24 & \textbf{98.41$\pm$0.05} & \textbf{98.23$\pm$0.69} & \textbf{97.10$\pm$0.27} & \underline{99.59$\pm$0.49} & 81.16$\pm$0.14 & \textbf{98.97$\pm$0.34} & \textbf{87.49$\pm$1.29} & \textbf{1.56} \\
\RepeatMixer & 99.04$\pm$0.01 & \textbf{99.15$\pm$0.01} & 95.36$\pm$0.49 & \underline{93.47$\pm$0.17} & 93.60$\pm$0.39 & 86.15$\pm$0.31 & 75.07$\pm$0.21 & 69.92$\pm$0.11 & 56.58$\pm$0.12 & 4.22 \\
\TIDFormer & \underline{99.19$\pm$0.09} & 98.93$\pm$0.16 & \underline{96.64$\pm$0.14} & 92.72$\pm$0.15 & 92.79$\pm$0.58 & \textbf{99.87$\pm$0.05} & 71.81$\pm$2.91 & 61.25$\pm$2.80 & 67.43$\pm$1.42 & 4.78 \\
\DyGFormer & 98.64$\pm$0.07 & 98.63$\pm$0.01 & 94.01$\pm$0.67 & 91.11$\pm$0.85 & 86.29$\pm$0.24 & 97.76$\pm$0.41 & 77.90$\pm$0.58 & 70.20$\pm$1.44 & 57.12$\pm$0.62 & 4.89 \\
\CNEN & 99.01$\pm$0.05 & \textbf{99.15$\pm$0.01} & 96.03$\pm$0.08 & 93.12$\pm$0.09 & \underline{95.69$\pm$0.07} & 89.71$\pm$0.07 & \textbf{84.14$\pm$0.63} & \underline{78.57$\pm$0.07} & 69.52$\pm$0.34 & \underline{2.56} \\
\TGAT & 96.67$\pm$0.07 & 98.47$\pm$0.02 & 78.53$\pm$0.74 & 68.89$\pm$1.10 & 87.11$\pm$0.19 & 75.69$\pm$0.78 & 75.84$\pm$1.99 & 64.01$\pm$0.12 & 52.83$\pm$1.12 & 8.89 \\
\TCL & 95.84$\pm$0.18 & 97.42$\pm$0.02 & 87.82$\pm$1.36 & 75.74$\pm$0.72 & 83.12$\pm$0.18 & 72.46$\pm$3.23 & 76.27$\pm$0.63 & 64.72$\pm$0.05 & 51.88$\pm$0.36 & 9.33 \\
\TGN & 98.37$\pm$0.07 & 98.60$\pm$0.06 & 92.03$\pm$1.13 & 88.32$\pm$0.99 & 91.21$\pm$1.15 & 76.99$\pm$1.80 & \underline{83.34$\pm$0.43} & 69.10$\pm$1.67 & \underline{69.71$\pm$2.65} & 5.44 \\
\CAWN & 98.54$\pm$0.04 & 99.01$\pm$0.01 & 93.87$\pm$0.08 & 90.45$\pm$0.14 & 80.38$\pm$0.26 & 75.70$\pm$3.27 & 77.16$\pm$0.39 & 68.54$\pm$0.18 & 53.09$\pm$0.22 & 6.56 \\
\EdgeBank & 90.78$\pm$0.00 & 95.37$\pm$0.00 & 77.30$\pm$0.00 & 87.05$\pm$0.00 & 60.86$\pm$0.00 & 64.14$\pm$0.00 & 62.57$\pm$0.00 & 66.75$\pm$0.00 & 62.97$\pm$0.00 & 9.56 \\
\GraphMixer & 96.92$\pm$0.03 & 97.17$\pm$0.02 & 91.81$\pm$0.67 & 84.38$\pm$0.21 & 84.01$\pm$0.17 & 83.17$\pm$0.53 & 76.96$\pm$0.79 & 65.52$\pm$0.51 & 52.46$\pm$0.27 & 8.11 \\
\bottomrule
\end{tabularx}%
}
\vspace{0.4em}
\textbf{Historical  Negative Sampling}\\[0.2em]
\resizebox{0.98\textwidth}{!}{%
\begin{tabularx}{\textwidth}{l*{9}{>{\centering\arraybackslash}X}c}
\toprule
Baseline & \Wikipedia & \Reddit & \UCI & \Enron & \Mooc & \CanParl & \USLegis & \UNTrade & \UNVote & AR \\
\midrule
\DiffDyG & \underline{86.34$\pm$0.02} & \textbf{82.16$\pm$0.07} & \textbf{84.19$\pm$0.63} & \underline{82.67$\pm$0.05} & \underline{92.02$\pm$0.46} & \textbf{97.83$\pm$0.06} & 80.21$\pm$0.17 & \underline{79.10$\pm$1.78} & \underline{81.58$\pm$1.21} & \textbf{2.11} \\
\RepeatMixer & 85.32$\pm$0.70 & 81.95$\pm$0.56 & 78.85$\pm$0.43 & \textbf{84.33$\pm$0.13} & \textbf{92.09$\pm$0.32} & 81.83$\pm$1.66 & 71.89$\pm$9.62 & 63.69$\pm$3.02 & 56.64$\pm$0.04 & \underline{4.22} \\
\TIDFormer & 83.14$\pm$0.23 & \underline{81.99$\pm$0.21} & \underline{84.19$\pm$0.77} & 78.78$\pm$2.94 & 88.55$\pm$0.31 & 68.96$\pm$1.28 & 68.23$\pm$0.58 & 59.90$\pm$2.30 & 61.16$\pm$1.66 & 5.78 \\
\DyGFormer & 78.80$\pm$1.95 & 80.54$\pm$0.29 & 76.97$\pm$0.24 & 76.55$\pm$0.52 & 87.04$\pm$0.35 & \underline{97.61$\pm$0.40} & \textbf{90.77$\pm$1.96} & 73.86$\pm$1.13 & 64.27$\pm$1.78 & 5.11 \\
\CNEN & 76.90$\pm$1.34 & 81.46$\pm$0.50 & 78.98$\pm$0.36 & 78.05$\pm$0.08 & 88.25$\pm$0.09 & 79.20$\pm$0.60 & 83.93$\pm$3.99 & 76.49$\pm$0.25 & 70.45$\pm$0.85 & 4.56 \\
\TGAT & 82.87$\pm$0.22 & 79.33$\pm$0.16 & 58.89$\pm$1.57 & 61.85$\pm$1.43 & 80.81$\pm$0.67 & 70.86$\pm$0.94 & 73.47$\pm$5.25 & 60.37$\pm$0.68 & 53.95$\pm$3.15 & 8.44 \\
\TCL & 85.76$\pm$0.46 & 76.49$\pm$0.16 & 72.25$\pm$3.46 & 67.95$\pm$0.88 & 72.09$\pm$0.56 & 69.95$\pm$3.70 & 83.97$\pm$3.71 & 61.43$\pm$1.04 & 52.29$\pm$2.39 & 7.89 \\
\TGN & 82.74$\pm$0.32 & 81.11$\pm$0.19 & 77.25$\pm$2.68 & 77.09$\pm$2.22 & 88.00$\pm$1.80 & 73.23$\pm$3.08 & 83.53$\pm$4.53 & 63.93$\pm$5.41 & 73.40$\pm$5.20 & 5.33 \\
\CAWN & 67.84$\pm$0.64 & 80.27$\pm$0.30 & 57.86$\pm$0.15 & 65.10$\pm$0.34 & 71.57$\pm$1.07 & 72.06$\pm$3.94 & 78.62$\pm$7.46 & 63.09$\pm$0.74 & 51.27$\pm$0.33 & 9.11 \\
\EdgeBank & 77.27$\pm$0.00 & 78.58$\pm$0.00 & 69.56$\pm$0.00 & 79.59$\pm$0.00 & 61.90$\pm$0.00 & 63.04$\pm$0.00 & 67.41$\pm$0.00 & \textbf{86.61$\pm$0.00} & \textbf{89.62$\pm$0.00} & 7.22 \\
\GraphMixer & \textbf{87.68$\pm$0.17} & 77.80$\pm$0.12 & 77.54$\pm$2.02 & 75.27$\pm$1.14 & 76.68$\pm$1.40 & 79.03$\pm$1.01 & \underline{85.17$\pm$0.70} & 63.20$\pm$1.54 & 52.61$\pm$1.44 & 6.11 \\
\bottomrule
\end{tabularx}%
}
\vspace{0.4em}
\textbf{Inductive  Negative Sampling}\\[0.2em]
\resizebox{0.98\textwidth}{!}{%
\begin{tabularx}{\textwidth}{l*{9}{>{\centering\arraybackslash}X}c}
\toprule
Baseline & \Wikipedia & \Reddit & \UCI & \Enron & \Mooc & \CanParl & \USLegis & \UNTrade & \UNVote & AR \\
\midrule
\DiffDyG & \underline{84.53$\pm$0.07} & \underline{88.47$\pm$3.15} & \textbf{83.13$\pm$0.28} & \textbf{80.82$\pm$0.03} & \textbf{85.31$\pm$0.22} & \textbf{96.73$\pm$0.14} & 79.97$\pm$3.02 & \textbf{77.80$\pm$0.51} & \textbf{80.62$\pm$0.14} & \textbf{1.67} \\
\RepeatMixer & 83.67$\pm$0.89 & 86.45$\pm$1.03 & 76.56$\pm$0.30 & \underline{80.05$\pm$0.53} & 81.94$\pm$0.95 & 82.46$\pm$1.62 & 76.01$\pm$0.61 & 63.69$\pm$3.02 & 56.64$\pm$0.04 & \underline{4.78} \\
\TIDFormer & \textbf{86.13$\pm$1.14} & \textbf{88.87$\pm$0.65} & \underline{82.95$\pm$0.38} & 78.30$\pm$2.08 & \underline{83.43$\pm$0.69} & 64.76$\pm$1.38 & 68.89$\pm$2.08 & 60.22$\pm$0.98 & 59.54$\pm$0.94 & 5.00 \\
\DyGFormer & 75.09$\pm$3.70 & 86.23$\pm$0.51 & 65.96$\pm$1.18 & 74.07$\pm$0.64 & 80.76$\pm$0.76 & \underline{96.70$\pm$0.59} & \textbf{87.96$\pm$1.80} & 62.56$\pm$1.51 & 53.37$\pm$1.26 & 5.78 \\
\CNEN & 70.22$\pm$1.47 & 85.66$\pm$0.34 & 65.63$\pm$0.77 & 74.34$\pm$0.18 & 79.79$\pm$0.26 & 78.29$\pm$1.03 & 80.13$\pm$4.77 & 73.77$\pm$0.26 & 68.38$\pm$0.68 & 5.78 \\
\TGAT & 81.93$\pm$0.22 & 87.13$\pm$0.20 & 60.80$\pm$1.01 & 60.45$\pm$2.12 & 73.18$\pm$0.33 & 72.47$\pm$1.18 & 71.62$\pm$5.42 & 66.13$\pm$0.78 & 53.04$\pm$2.58 & 7.56 \\
\TCL & 82.19$\pm$0.48 & 84.67$\pm$0.29 & 70.05$\pm$1.86 & 67.64$\pm$0.86 & 70.36$\pm$0.37 & 69.47$\pm$2.12 & 82.54$\pm$3.91 & 67.80$\pm$1.21 & 52.02$\pm$1.64 & 7.22 \\
\TGN & 80.97$\pm$0.31 & 84.56$\pm$0.24 & 64.11$\pm$1.04 & 71.34$\pm$2.46 & 77.44$\pm$2.86 & 69.57$\pm$2.81 & 78.12$\pm$4.46 & 66.37$\pm$5.39 & 72.69$\pm$3.72 & 7.22 \\
\CAWN & 70.95$\pm$0.95 & 88.04$\pm$0.29 & 58.06$\pm$0.26 & 75.17$\pm$0.50 & 70.32$\pm$1.43 & 72.93$\pm$1.78 & 76.45$\pm$7.02 & 71.73$\pm$0.74 & 52.75$\pm$0.90 & 6.89 \\
\EdgeBank & 81.73$\pm$0.00 & 85.93$\pm$0.00 & 58.03$\pm$0.00 & 75.00$\pm$0.00 & 48.18$\pm$0.00 & 61.41$\pm$0.00 & 68.66$\pm$0.00 & \underline{74.20$\pm$0.00} & \underline{72.85$\pm$0.00} & 7.44 \\
\GraphMixer & 84.28$\pm$0.30 & 82.21$\pm$0.13 & 74.59$\pm$0.74 & 71.53$\pm$0.85 & 72.45$\pm$0.72 & 70.52$\pm$0.94 & \underline{84.22$\pm$0.91} & 66.53$\pm$1.22 & 51.89$\pm$0.74 & 6.67 \\
\bottomrule
\end{tabularx}%
}
\end{table*}
\begin{table*}[!t]
\caption{\small Performance (Average Precision) comparison in the inductive setting. Average Rank (AR) is computed across the 9 datasets. The best and second best results are shown in \textbf{bold} and  \underline{underlined}.}
\label{tab:ap_inductive_nss}
\centering
\scriptsize
\setlength{\tabcolsep}{2pt}
\textbf{Random Negative Sampling}\\[0.2em]
\resizebox{0.98\textwidth}{!}{%
\begin{tabularx}{\textwidth}{l*{9}{>{\centering\arraybackslash}X}c}
\toprule
Baseline & \Wikipedia & \Reddit & \UCI & \Enron & \Mooc & \CanParl & \USLegis & \UNTrade & \UNVote & AR \\
\midrule
\DiffDyG & \textbf{98.97$\pm$0.25} & \textbf{99.04$\pm$0.01} & \textbf{98.06$\pm$0.37} & \textbf{94.48$\pm$1.95} & \textbf{95.16$\pm$0.70} & \textbf{98.12$\pm$0.34} & \textbf{73.78$\pm$0.39} & \textbf{97.82$\pm$1.00} & \textbf{74.46$\pm$1.82} & \textbf{1.00} \\
\RepeatMixer & 98.70$\pm$0.05 & 98.85$\pm$0.01 & 95.04$\pm$0.12 & 87.97$\pm$0.29 & 93.05$\pm$0.12 & 59.77$\pm$0.70 & 51.70$\pm$0.97 & 61.36$\pm$0.21 & 54.47$\pm$0.08 & 4.78 \\
\TIDFormer & \underline{98.93$\pm$0.02} & \underline{98.98$\pm$0.03} & \underline{95.95$\pm$0.12} & \underline{89.92$\pm$0.45} & \underline{93.29$\pm$0.39} & \underline{93.97$\pm$2.93} & \underline{60.83$\pm$1.25} & 60.31$\pm$1.87 & 66.02$\pm$1.34 & \underline{2.89} \\
\DyGFormer & 98.36$\pm$0.03 & 98.35$\pm$0.02 & 94.14$\pm$0.43 & 88.63$\pm$0.18 & 85.27$\pm$0.09 & 87.74$\pm$0.71 & 54.28$\pm$2.87 & 64.55$\pm$0.62 & 55.93$\pm$0.39 & 4.89 \\
\CNEN & 98.37$\pm$0.03 & 98.78$\pm$0.01 & 95.03$\pm$0.16 & 89.66$\pm$0.22 & 91.89$\pm$0.31 & 68.31$\pm$0.59 & 59.44$\pm$0.44 & \underline{66.58$\pm$0.27} & \underline{69.71$\pm$0.48} & 3.33 \\
\TGAT & 96.22$\pm$0.07 & 97.09$\pm$0.04 & 79.54$\pm$0.48 & 67.05$\pm$1.51 & 85.50$\pm$0.19 & 55.18$\pm$0.79 & 51.00$\pm$3.11 & 61.03$\pm$0.18 & 52.24$\pm$1.46 & 8.33 \\
\TCL & 96.22$\pm$0.17 & 94.09$\pm$0.07 & 87.36$\pm$2.03 & 76.14$\pm$0.79 & 80.60$\pm$0.22 & 54.30$\pm$0.66 & 52.59$\pm$0.97 & 62.21$\pm$0.12 & 51.60$\pm$0.97 & 8.44 \\
\TGN & 97.83$\pm$0.04 & 97.50$\pm$0.07 & 88.12$\pm$2.05 & 77.94$\pm$1.02 & 89.04$\pm$1.17 & 54.10$\pm$0.93 & 58.63$\pm$0.37 & 58.31$\pm$3.15 & 58.85$\pm$2.51 & 6.89 \\
\CAWN & 98.24$\pm$0.03 & 98.62$\pm$0.01 & 92.73$\pm$0.06 & 86.35$\pm$0.51 & 81.42$\pm$0.24 & 55.80$\pm$0.69 & 53.17$\pm$1.20 & 65.24$\pm$0.21 & 49.94$\pm$0.45 & 6.33 \\
\GraphMixer & 96.65$\pm$0.02 & 95.26$\pm$0.02 & 91.19$\pm$0.42 & 75.88$\pm$0.48 & 81.41$\pm$0.21 & 55.91$\pm$0.82 & 50.71$\pm$0.76 & 62.17$\pm$0.31 & 50.68$\pm$0.44 & 8.11 \\
\bottomrule
\end{tabularx}%
}
\vspace{0.4em}
\textbf{Historical Negative Sampling}\\[0.2em]
\resizebox{0.98\textwidth}{!}{%
\begin{tabularx}{\textwidth}{l*{9}{>{\centering\arraybackslash}X}c}
\toprule
Baseline & \Wikipedia & \Reddit & \UCI & \Enron & \Mooc & \CanParl & \USLegis & \UNTrade & \UNVote & AR \\
\midrule
\DiffDyG & \underline{85.67$\pm$1.12} & \underline{66.08$\pm$0.05} & \textbf{86.27$\pm$2.14} & 74.32$\pm$0.13 & \underline{84.37$\pm$0.06} & \textbf{88.84$\pm$0.31} & \textbf{69.88$\pm$0.81} & \textbf{67.24$\pm$4.14} & \textbf{67.35$\pm$2.95} & \textbf{1.56} \\
\RepeatMixer & 84.11$\pm$1.88 & \textbf{66.14$\pm$1.40} & 85.52$\pm$0.16 & \textbf{82.86$\pm$0.47} & 83.19$\pm$1.72 & 56.14$\pm$1.34 & 53.90$\pm$2.90 & 57.48$\pm$1.95 & 54.53$\pm$0.13 & 4.00 \\
\TIDFormer & 84.99$\pm$1.89 & 65.62$\pm$0.92 & \underline{86.11$\pm$1.81} & \underline{76.99$\pm$1.21} & \textbf{84.61$\pm$1.17} & 63.84$\pm$0.56 & \underline{69.56$\pm$2.11} & 59.13$\pm$1.41 & 60.15$\pm$0.26 & \underline{2.67} \\
\DyGFormer & 71.42$\pm$4.43 & 65.37$\pm$0.60 & 72.13$\pm$1.87 & 67.07$\pm$0.62 & 80.82$\pm$0.30 & \underline{87.40$\pm$0.85} & 56.31$\pm$3.46 & 53.20$\pm$1.07 & 52.63$\pm$1.26 & 5.89 \\
\CNEN & 70.70$\pm$2.75 & 64.09$\pm$0.92 & 70.23$\pm$1.75 & 70.90$\pm$0.39 & 80.59$\pm$0.10 & 66.14$\pm$0.77 & 63.48$\pm$1.31 & \underline{60.48$\pm$0.15} & \underline{64.39$\pm$0.28} & 5.00 \\
\TGAT & 84.17$\pm$0.22 & 63.47$\pm$0.36 & 70.52$\pm$0.93 & 61.40$\pm$1.31 & 76.73$\pm$0.29 & 56.72$\pm$0.47 & 51.83$\pm$3.95 & 55.28$\pm$0.71 & 53.05$\pm$3.10 & 7.33 \\
\TCL & 82.20$\pm$2.18 & 60.83$\pm$0.25 & 76.71$\pm$1.00 & 67.11$\pm$0.62 & 74.27$\pm$0.53 & 55.71$\pm$0.74 & 53.87$\pm$1.41 & 55.76$\pm$1.03 & 54.19$\pm$2.17 & 7.11 \\
\TGN & 81.76$\pm$0.32 & 64.85$\pm$0.85 & 70.78$\pm$0.78 & 62.91$\pm$1.16 & 77.07$\pm$3.41 & 54.42$\pm$0.77 & 61.18$\pm$1.10 & 52.80$\pm$3.19 & 63.74$\pm$3.00 & 6.67 \\
\CAWN & 67.27$\pm$1.63 & 63.67$\pm$0.41 & 64.54$\pm$0.47 & 60.70$\pm$0.36 & 74.68$\pm$0.68 & 57.14$\pm$0.07 & 55.56$\pm$1.71 & 55.00$\pm$0.38 & 47.98$\pm$0.84 & 8.22 \\
\GraphMixer & \textbf{87.60$\pm$0.30} & 64.50$\pm$0.26 & 81.66$\pm$0.49 & 72.37$\pm$1.37 & 74.00$\pm$0.97 & 55.84$\pm$0.73 & 52.03$\pm$1.02 & 54.94$\pm$0.97 & 48.09$\pm$0.43 & 6.56 \\
\bottomrule
\end{tabularx}%
}
\vspace{0.4em}
\textbf{Inductive Negative Sampling}\\[0.2em]
\resizebox{0.98\textwidth}{!}{%
\begin{tabularx}{\textwidth}{l*{9}{>{\centering\arraybackslash}X}c}
\toprule
Baseline & \Wikipedia & \Reddit & \UCI & \Enron & \Mooc & \CanParl & \USLegis & \UNTrade & \UNVote & AR \\
\midrule
\DiffDyG & \underline{85.67$\pm$1.12} & \underline{66.07$\pm$0.05} & \textbf{86.27$\pm$2.14} & 74.33$\pm$0.13 & \underline{84.36$\pm$0.06} & \textbf{88.52$\pm$0.31} & \textbf{69.88$\pm$0.81} & \textbf{61.25$\pm$0.93} & \textbf{67.36$\pm$2.95} & \textbf{1.56} \\
\RepeatMixer & 84.10$\pm$1.88 & \textbf{66.13$\pm$1.40} & 85.53$\pm$0.16 & \textbf{82.87$\pm$0.47} & 83.19$\pm$1.72 & 56.92$\pm$1.36 & 52.18$\pm$1.53 & 57.48$\pm$1.95 & 54.53$\pm$0.13 & 4.00 \\
\TIDFormer & 84.99$\pm$1.89 & 65.62$\pm$0.92 & \underline{86.11$\pm$1.81} & \underline{76.98$\pm$1.21} & \textbf{84.61$\pm$1.17} & 63.27$\pm$0.32 & \underline{69.56$\pm$2.11} & 59.13$\pm$1.41 & 59.09$\pm$0.19 & \underline{2.67} \\
\DyGFormer & 71.42$\pm$4.43 & 65.35$\pm$0.60 & 72.13$\pm$1.86 & 67.07$\pm$0.62 & 80.82$\pm$0.30 & \underline{87.22$\pm$0.82} & 56.31$\pm$3.46 & 52.56$\pm$1.70 & 52.61$\pm$1.25 & 6.00 \\
\CNEN & 70.73$\pm$2.76 & 64.08$\pm$0.91 & 70.24$\pm$1.74 & 71.03$\pm$0.47 & 80.58$\pm$0.11 & 65.83$\pm$0.62 & 63.94$\pm$0.97 & \underline{60.53$\pm$0.17} & \underline{64.72$\pm$0.35} & 5.00 \\
\TGAT & 84.17$\pm$0.22 & 63.40$\pm$0.36 & 70.49$\pm$0.93 & 61.40$\pm$1.30 & 76.72$\pm$0.30 & 56.46$\pm$0.50 & 51.83$\pm$3.95 & 55.58$\pm$0.68 & 53.08$\pm$3.10 & 7.44 \\
\TCL & 82.20$\pm$2.18 & 60.81$\pm$0.26 & 76.65$\pm$0.99 & 67.11$\pm$0.62 & 74.28$\pm$0.53 & 55.46$\pm$0.69 & 53.87$\pm$1.41 & 55.66$\pm$0.98 & 54.13$\pm$2.16 & 7.00 \\
\TGN & 81.77$\pm$0.32 & 64.84$\pm$0.84 & 70.73$\pm$0.79 & 62.90$\pm$1.16 & 77.07$\pm$3.40 & 54.18$\pm$0.73 & 61.18$\pm$1.10 & 52.80$\pm$3.24 & 63.71$\pm$2.97 & 6.56 \\
\CAWN & 67.24$\pm$1.63 & 63.65$\pm$0.41 & 64.54$\pm$0.47 & 60.72$\pm$0.36 & 74.69$\pm$0.68 & 57.06$\pm$0.08 & 55.56$\pm$1.71 & 54.97$\pm$0.38 & 48.01$\pm$0.82 & 8.22 \\
\GraphMixer & \textbf{87.60$\pm$0.29} & 64.49$\pm$0.25 & 81.64$\pm$0.49 & 72.37$\pm$1.38 & 73.99$\pm$0.97 & 55.76$\pm$0.65 & 52.03$\pm$1.02 & 54.88$\pm$1.01 & 48.10$\pm$0.40 & 6.56 \\
\bottomrule
\end{tabularx}%
}
\end{table*}
\begin{table*}[!t]
\caption{\small Performance (AUC-ROC) comparison in the inductive setting. Average Rank (AR) is computed across the 9 datasets. The best and second best results are shown in \textbf{bold} and  \underline{underlined}.}
\label{tab:auc_inductive_nss}
\centering
\scriptsize
\setlength{\tabcolsep}{2pt}
\textbf{Random Negative Sampling}\\[0.2em]
\resizebox{0.98\textwidth}{!}{%
\begin{tabularx}{\textwidth}{l*{9}{>{\centering\arraybackslash}X}c}
\toprule
Baseline & \Wikipedia & \Reddit & \UCI & \Enron & \Mooc & \CanParl & \USLegis & \UNTrade & \UNVote & AR \\
\midrule
\DiffDyG & \textbf{98.97$\pm$0.13} & \textbf{98.79$\pm$0.62} & \textbf{97.32$\pm$0.54} & \textbf{92.59$\pm$0.47} & \textbf{95.40$\pm$0.90} & \textbf{98.05$\pm$0.64} & 59.38$\pm$1.64 & \textbf{98.81$\pm$0.03} & \textbf{70.99$\pm$1.16} & \textbf{1.33} \\
\RepeatMixer & 98.63$\pm$0.02 & \underline{98.69$\pm$0.02} & \underline{93.54$\pm$0.14} & 89.04$\pm$0.26 & \underline{94.10$\pm$0.07} & 60.03$\pm$1.19 & 49.29$\pm$1.90 & 66.17$\pm$0.17 & 54.32$\pm$0.25 & 4.00 \\
\TIDFormer & \underline{98.76$\pm$0.09} & 98.23$\pm$0.28 & 93.35$\pm$0.13 & 85.62$\pm$0.31 & 92.55$\pm$0.58 & \underline{94.52$\pm$2.69} & 59.55$\pm$2.87 & 61.14$\pm$2.84 & 65.13$\pm$1.00 & 4.11 \\
\DyGFormer & 98.16$\pm$0.32 & 98.05$\pm$0.03 & 91.97$\pm$1.68 & 88.56$\pm$1.07 & 85.00$\pm$0.10 & 89.33$\pm$0.48 & 53.21$\pm$3.04 & 67.25$\pm$1.05 & 56.73$\pm$0.69 & 4.78 \\
\CNEN & 98.23$\pm$0.01 & 98.62$\pm$0.01 & 93.34$\pm$0.17 & \underline{90.18$\pm$0.15} & 92.76$\pm$0.29 & 70.22$\pm$0.77 & \underline{61.52$\pm$0.52} & \underline{67.26$\pm$0.10} & \underline{69.17$\pm$0.53} & \underline{2.89} \\
\TGAT & 95.90$\pm$0.09 & 96.98$\pm$0.04 & 77.64$\pm$0.38 & 64.63$\pm$1.74 & 86.84$\pm$0.17 & 56.51$\pm$0.75 & 48.27$\pm$3.50 & 62.72$\pm$0.12 & 51.83$\pm$1.35 & 8.33 \\
\TCL & 95.57$\pm$0.20 & 93.80$\pm$0.07 & 84.49$\pm$1.82 & 72.33$\pm$0.99 & 81.43$\pm$0.19 & 55.83$\pm$1.07 & 50.43$\pm$1.48 & 63.76$\pm$0.07 & 50.51$\pm$1.05 & 8.78 \\
\TGN & 97.72$\pm$0.03 & 97.39$\pm$0.07 & 86.68$\pm$2.29 & 78.83$\pm$1.11 & 91.24$\pm$0.99 & 55.86$\pm$0.75 & \textbf{62.38$\pm$0.48} & 59.99$\pm$3.50 & 61.23$\pm$2.71 & 6.44 \\
\CAWN & 98.03$\pm$0.04 & 98.42$\pm$0.02 & 90.40$\pm$0.11 & 87.02$\pm$0.50 & 81.86$\pm$0.25 & 58.83$\pm$1.13 & 51.49$\pm$1.13 & 67.05$\pm$0.21 & 48.34$\pm$0.76 & 6.22 \\
\GraphMixer & 96.30$\pm$0.04 & 94.97$\pm$0.05 & 89.30$\pm$0.57 & 76.51$\pm$0.71 & 82.77$\pm$0.24 & 58.32$\pm$1.08 & 47.20$\pm$0.89 & 63.48$\pm$0.37 & 50.04$\pm$0.86 & 8.11 \\
\bottomrule
\end{tabularx}%
}
\vspace{0.4em}
\textbf{Historical Negative Sampling}\\[0.2em]
\resizebox{0.98\textwidth}{!}{%
\begin{tabularx}{\textwidth}{l*{9}{>{\centering\arraybackslash}X}c}
\toprule
Baseline & \Wikipedia & \Reddit & \UCI & \Enron & \Mooc & \CanParl & \USLegis & \UNTrade & \UNVote & AR \\
\midrule
\DiffDyG & \underline{80.19$\pm$0.93} & \textbf{65.29$\pm$1.58} & \textbf{78.02$\pm$0.34} & \underline{74.14$\pm$0.01} & \textbf{82.79$\pm$1.51} & \textbf{89.18$\pm$1.78} & 55.16$\pm$0.37 & \textbf{67.87$\pm$2.31} & \underline{64.00$\pm$3.15} & \textbf{1.78} \\
\RepeatMixer & 77.58$\pm$1.00 & \underline{65.15$\pm$0.36} & 76.91$\pm$0.23 & \textbf{78.84$\pm$0.29} & \underline{82.68$\pm$1.28} & 54.27$\pm$2.50 & 53.26$\pm$3.85 & \underline{61.67$\pm$2.81} & 54.62$\pm$0.18 & \underline{4.22} \\
\TIDFormer & 73.81$\pm$2.16 & 64.87$\pm$1.13 & \underline{77.35$\pm$0.57} & 72.86$\pm$2.45 & 79.63$\pm$1.29 & 64.82$\pm$0.74 & \textbf{67.90$\pm$0.91} & 59.03$\pm$2.74 & 59.05$\pm$0.33 & \underline{4.22} \\
\DyGFormer & 68.33$\pm$2.82 & 64.81$\pm$0.25 & 65.55$\pm$1.01 & 65.78$\pm$0.42 & 80.77$\pm$0.63 & \underline{88.68$\pm$0.74} & 56.57$\pm$3.22 & 58.46$\pm$1.65 & 53.85$\pm$2.02 & 5.89 \\
\CNEN & 69.88$\pm$1.81 & 63.18$\pm$0.42 & 67.78$\pm$0.66 & 70.57$\pm$0.25 & 80.83$\pm$0.14 & 67.25$\pm$0.99 & 64.26$\pm$1.02 & 60.95$\pm$0.23 & 63.86$\pm$0.68 & 4.78 \\
\TGAT & 78.38$\pm$0.20 & 64.43$\pm$0.27 & 62.32$\pm$1.18 & 57.84$\pm$2.18 & 74.08$\pm$0.27 & 58.30$\pm$0.61 & 49.99$\pm$4.88 & 59.74$\pm$0.59 & 51.73$\pm$4.12 & 7.44 \\
\TCL & 79.79$\pm$0.96 & 61.43$\pm$0.26 & 70.46$\pm$1.94 & 64.06$\pm$1.02 & 69.82$\pm$0.32 & 57.30$\pm$1.03 & 52.12$\pm$2.13 & 61.12$\pm$0.97 & 54.66$\pm$2.11 & 6.44 \\
\TGN & 75.75$\pm$0.29 & 64.55$\pm$0.50 & 62.69$\pm$0.90 & 62.68$\pm$1.09 & 77.69$\pm$3.55 & 55.64$\pm$0.54 & \underline{64.87$\pm$1.65} & 55.61$\pm$3.54 & \textbf{68.59$\pm$3.11} & 6.22 \\
\CAWN & 62.04$\pm$0.65 & 64.94$\pm$0.21 & 56.39$\pm$0.10 & 62.25$\pm$0.40 & 71.68$\pm$0.94 & 60.11$\pm$0.48 & 54.41$\pm$1.31 & 60.95$\pm$0.80 & 48.01$\pm$1.77 & 7.22 \\
\GraphMixer & \textbf{82.87$\pm$0.21} & 64.27$\pm$0.13 & 75.98$\pm$0.84 & 68.20$\pm$1.62 & 72.53$\pm$0.84 & 56.68$\pm$1.20 & 49.28$\pm$0.86 & 59.88$\pm$1.17 & 45.49$\pm$0.42 & 6.67 \\
\bottomrule
\end{tabularx}%
}
\vspace{0.4em}
\textbf{Inductive Negative Sampling}\\[0.2em]
\resizebox{0.98\textwidth}{!}{%
\begin{tabularx}{\textwidth}{l*{9}{>{\centering\arraybackslash}X}c}
\toprule
Baseline & \Wikipedia & \Reddit & \UCI & \Enron & \Mooc & \CanParl & \USLegis & \UNTrade & \UNVote & AR \\
\midrule
\DiffDyG & 80.19$\pm$0.93 & \textbf{65.29$\pm$1.58} & \textbf{78.02$\pm$0.34} & \underline{74.14$\pm$0.01} & \textbf{82.79$\pm$1.54} & \textbf{88.76$\pm$1.78} & 55.16$\pm$0.37 & \textbf{64.07$\pm$4.17} & 64.02$\pm$3.15 & \textbf{2.00} \\
\RepeatMixer & 77.58$\pm$1.00 & \underline{65.15$\pm$0.36} & 76.92$\pm$0.23 & \textbf{78.85$\pm$0.29} & \underline{82.68$\pm$1.28} & 55.68$\pm$2.54 & 51.96$\pm$1.70 & \underline{61.67$\pm$2.81} & 54.62$\pm$0.18 & 4.33 \\
\TIDFormer & \underline{80.38$\pm$0.10} & 64.87$\pm$1.15 & \underline{77.35$\pm$0.57} & 72.86$\pm$2.45 & 79.63$\pm$1.29 & 64.24$\pm$0.50 & \textbf{67.90$\pm$0.16} & 59.03$\pm$2.74 & 59.05$\pm$0.33 & \underline{3.67} \\
\DyGFormer & 68.33$\pm$2.82 & 64.80$\pm$0.25 & 65.58$\pm$1.00 & 65.79$\pm$0.42 & 80.77$\pm$0.63 & \underline{88.51$\pm$0.73} & 56.57$\pm$3.22 & 57.28$\pm$3.06 & 53.87$\pm$2.01 & 5.89 \\
\CNEN & 69.89$\pm$1.86 & 63.17$\pm$0.48 & 67.80$\pm$0.64 & 70.63$\pm$0.27 & 80.82$\pm$0.15 & 66.91$\pm$1.13 & 64.85$\pm$0.81 & 61.01$\pm$0.16 & \underline{64.36$\pm$0.45} & 4.56 \\
\TGAT & 78.38$\pm$0.20 & 64.39$\pm$0.27 & 62.29$\pm$1.17 & 57.83$\pm$2.18 & 74.07$\pm$0.27 & 58.15$\pm$0.62 & 49.99$\pm$4.88 & 59.98$\pm$0.59 & 51.78$\pm$4.14 & 7.44 \\
\TCL & 79.79$\pm$0.96 & 61.36$\pm$0.26 & 70.42$\pm$1.93 & 64.05$\pm$1.02 & 69.83$\pm$0.32 & 56.88$\pm$0.93 & 52.12$\pm$2.13 & 61.01$\pm$0.93 & 54.65$\pm$2.20 & 6.44 \\
\TGN & 75.76$\pm$0.29 & 64.55$\pm$0.50 & 62.66$\pm$0.91 & 62.68$\pm$1.09 & 77.68$\pm$3.55 & 55.43$\pm$0.42 & \underline{64.87$\pm$1.65} & 55.62$\pm$3.59 & \textbf{68.58$\pm$3.08} & 6.44 \\
\CAWN & 62.02$\pm$0.65 & 64.91$\pm$0.21 & 56.39$\pm$0.11 & 62.27$\pm$0.40 & 71.69$\pm$0.94 & 60.01$\pm$0.47 & 54.41$\pm$1.31 & 60.88$\pm$0.79 & 48.04$\pm$1.76 & 7.33 \\
\GraphMixer & \textbf{82.88$\pm$0.21} & 64.27$\pm$0.13 & 75.97$\pm$0.85 & 68.19$\pm$1.63 & 72.52$\pm$0.84 & 56.63$\pm$1.09 & 49.28$\pm$0.86 & 59.71$\pm$1.17 & 45.57$\pm$0.41 & 6.78 \\
\bottomrule
\end{tabularx}%
}
\end{table*}

\section{Additional experimental results}
\label{app:additional_exp}

\subsection{Additional results on dynamic node classification}
\label{app:additional_exp:node_classification}

In addition to dynamic link prediction, we further evaluate \DiffDyG on dynamic node classification to examine whether the proposed model remains effective for node-level temporal prediction tasks. 
Following the experimental setup of \DyGFormer~\cite{yu2023towards}, we conduct experiments on the standard Wikipedia and Reddit benchmarks and report AUC-ROC.  

\begin{table}[!t]
\caption{\small Performance comparison on dynamic node classification. Results are reported in AUC-ROC. The best and second best results are shown in \textbf{bold} and \underline{underlined}.}
\label{tab:node_classification}
\centering
\scriptsize
\setlength{\tabcolsep}{6pt}
\begin{tabular}{lccc}
\toprule
Model & Wikipedia & Reddit & AR \\
\midrule
DiffDyG & \underline{88.12$\pm$0.73} & \textbf{70.94$\pm$0.65} & \textbf{1.5} \\
DyGFormer & 87.44$\pm$1.08 & 68.00$\pm$1.74 & 5.0 \\
TIDFormer & 87.53$\pm$1.12 & 69.59$\pm$1.70 & \underline{3.0} \\
TCL & 77.83$\pm$2.13 & 68.87$\pm$2.15 & 7.0 \\
CNEN & \textbf{88.37$\pm$0.28} & 66.84$\pm$1.19 & 4.0 \\
RepeatMixer & 80.94$\pm$0.19 & 68.12$\pm$0.64 & 7.0 \\
TGAT & 84.09$\pm$1.27 & \underline{70.04$\pm$1.09} & 5.0 \\
TGN & 86.38$\pm$2.34 & 63.27$\pm$0.90 & 8.0 \\
CAWN & 84.88$\pm$1.33 & 66.34$\pm$1.78 & 7.5 \\
GraphMixer & 86.80$\pm$0.79 & 64.22$\pm$3.32 & 7.0 \\
\bottomrule
\end{tabular}
\end{table}

As shown in Tab.~\ref{tab:node_classification}, \DiffDyG achieves the best result on Reddit and the second-best result on Wikipedia. Although \CNEN obtains a slightly higher AUC-ROC on Wikipedia, its performance drops on Reddit, while \DiffDyG remains consistently strong on both datasets. As a result, \DiffDyG achieves the best overall AR of 1.5 among all compared methods. These results indicate that the proposed model is not limited to dynamic link prediction, but also generalizes well to node-level temporal prediction tasks.

\subsection{Comparison result with different metric in both transductive and inductive settings}
\label{app:additional_exp:sampling}

In the main paper, Tab.~\ref{tab:ap_transductive_rand} reports the AP results in the transductive setting under random negative sampling strategies. Here, Tabs.~\ref{tab:ap_transductive_nss}, ~\ref{tab:auc_transductive_nss}, \ref{tab:ap_inductive_nss}, and \ref{tab:auc_inductive_nss} provide the results under all 3 negative sampling strategies for transductive AP, AUC-ROC, inductive AP, and inductive AUC-ROC, respectively.

The results show that DiffDyG consistently achieves the best overall ranking across different evaluation settings. For transductive AUC-ROC, DiffDyG obtains the best AR under random, historical, and inductive negative sampling, with AR values of 1.56, 2.11, and 1.67, respectively. This indicates that the advantage of DiffDyG is not limited to AP, but also holds under a threshold-independent ranking metric.

In the inductive setting, DiffDyG remains the strongest overall method. For AP, it achieves the best AR under all three negative sampling strategies, with AR values of 1.00, 1.56, and 1.56 under random, historical, and inductive negative sampling, respectively. For AUC-ROC, \DiffDyG again obtains the best AR under all three strategies, with AR values of 1.33, 1.78, and 2.00. These results are notable because the inductive setting is more challenging, as the model must generalize to unseen nodes rather than only predicting future links among previously observed nodes.

Overall, the consistent top rankings across AP and AUC-ROC, across transductive and inductive settings, and across three negative sampling strategies demonstrate that the gains of DiffDyG are robust. This supports the central claim that sharpening temporal attention helps the model identify informative historical interactions under different evaluation protocols, rather than overfitting to a particular choice of negative samples.

\begin{figure}[!t]
  \centering
  \begin{subfigure}[b]{0.45\linewidth}
    \centering
    \includegraphics[width=\linewidth, trim=0 0 0 220, clip]{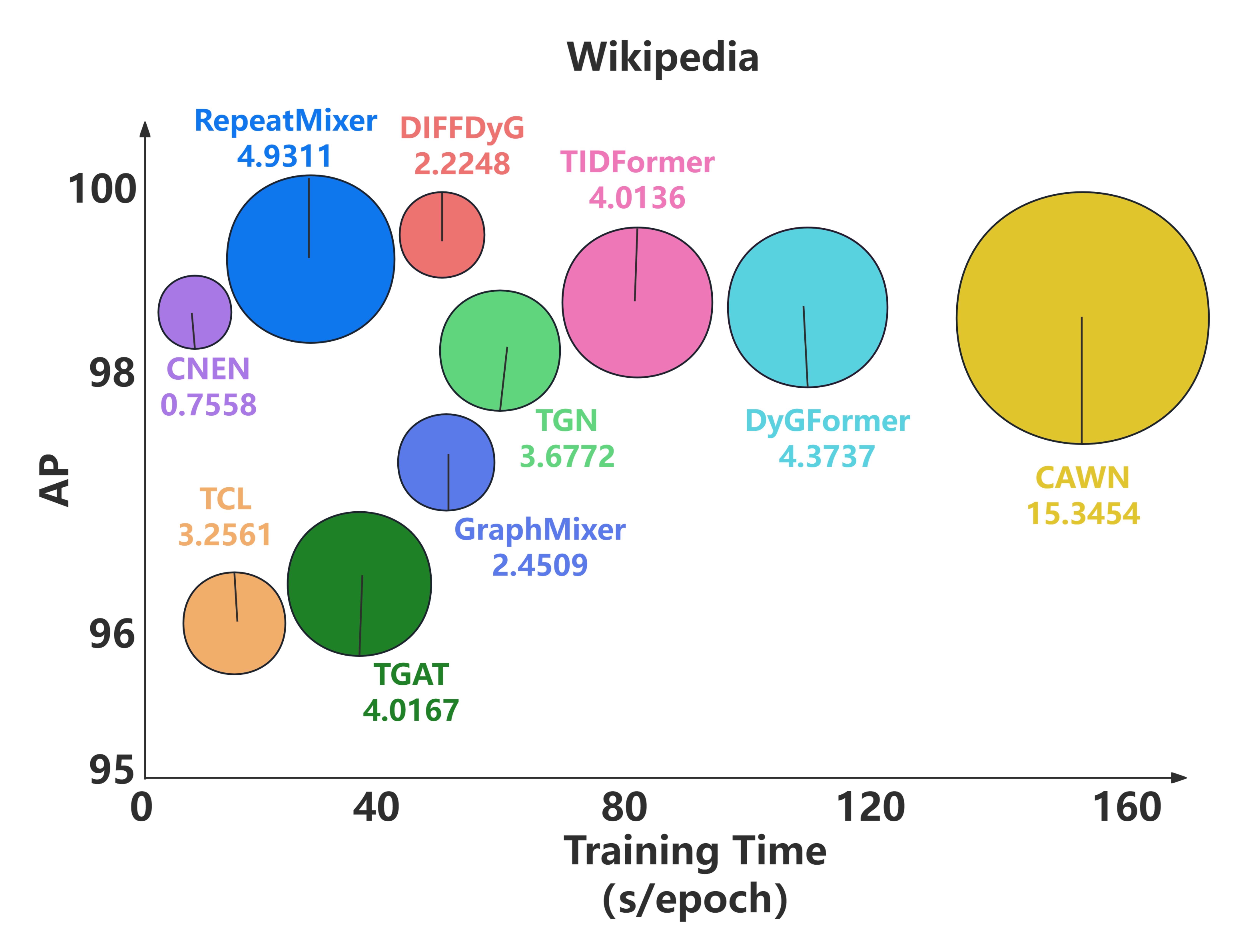}
    \caption{\Wikipedia}\label{fig:time_and_para_wiki}
  \end{subfigure}
  \begin{subfigure}[b]{0.45\linewidth}
    \centering
    \includegraphics[width=\linewidth, trim=0 0 0 220, clip]{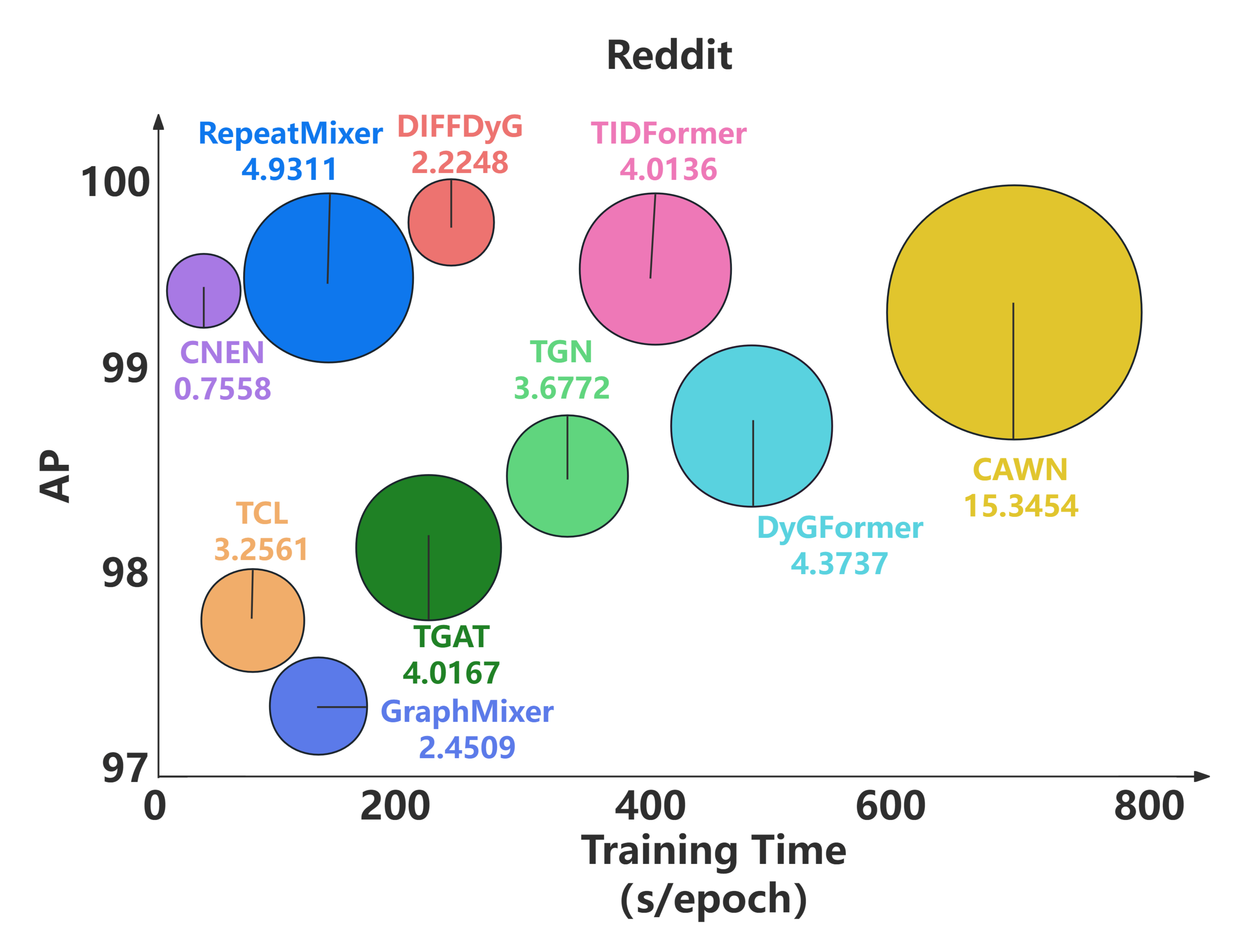}
    \caption{\Reddit}\label{fig:time_and_para_redd}
  \end{subfigure}
  \caption{Comparison of models' AP, parameter size (MB), and training time (seconds per epoch).}
  \label{fig:time_and_para}
\end{figure}

\subsection{Efficiency and model size}\label{sec:exp_efficiency}

Although differential attention introduces an additional softmax computation compared with standard attention, \DiffDyG remains efficient in practice. Its stronger temporal denoising allows the use of a smaller embedding dimension—36 in \DiffDyG versus 50 in \DyGFormer—and consequently fewer trainable parameters: $0.58$M for \DiffDyG, compared with $0.85$M for \TCL, $1.05$M for \TIDFormer, and $1.15$M for \DyGFormer. The gains in~\cref{sec:exp_main} are therefore not attributable to additional model capacity.

Combined with FlashAttention~\cite{dao2022flashattention} and Grouped Query Attention~\cite{ainslie2023gqa}, \DiffDyG trains $1.5\times$ to $2.5\times$ faster than \DyGFormer and \TIDFormer (Fig.~\ref{fig:time_and_para}), making it practical for large-scale dynamic graph learning while improving on both effectiveness and parameter efficiency.

\subsection{Hyperparameter sensitivity}\label{app:exp_hp_sensitivity}

We evaluate the sensitivity of \DiffDyG to five hyperparameters: the number of Transformer layers, attention heads, sampling hops, and the number of neighbors for both first and second hops. As shown in Tabs.~\ref{tab:hyperparameter_trans} and \ref{tab:hyperparameter_ind}, \DiffDyG demonstrates strong stability across different configurations. Performance remains largely consistent as the number of layers and attention heads increases, suggesting that the model is robust to variations in architectural depth and breadth.

When it comes to the neighbor sampling strategy, increasing the interaction range from 1-hop to 2-hop provides a noticeable performance improvement, as the 2-hop neighborhood introduces more topological context for representation learning. However, it is important to note that the performance drop from reducing to 1-hop is relatively minor compared to the impact of removing key architectural components, such as Differential Attention, RoPE, or Spatial Encoding, which were analyzed in the ablation studies in~\cref{sec:exp_ablation}.

\begin{table}[h!t]
\caption{Hyperparameter analysis in transductive setting}\label{tab:hyperparameter_trans}
\centering
\scriptsize
\renewcommand{\arraystretch}{1.1}
\begin{tabular}{l r | c c c c c c}
\toprule
\multicolumn{2}{c|}{Hyperparameter \& Value}  & \multicolumn{2}{c}{\Wikipedia} & \multicolumn{2}{c}{\UCI} & \multicolumn{2}{c}{\CanParl} \\
 & & AP & AUC-ROC & AP & AUC-ROC & AP & AUC-ROC \\
\midrule
\multirow{3}{*}{\# layers} 
    & 2 & 99.57$\pm$0.03 & 99.37$\pm$0.38 & 98.54$\pm$0.03 & 98.41$\pm$0.05 & 99.63$\pm$0.09 & 99.59$\pm$0.49 \\
    & 4 & 99.61$\pm$0.07 & 99.42$\pm$0.05 & 98.46$\pm$0.52 & 97.56$\pm$0.47 & 99.74$\pm$0.06 & 99.63$\pm$0.08 \\
    & 6 & 99.67$\pm$0.12 & 99.53$\pm$0.28 & 98.82$\pm$0.05 & 98.52$\pm$0.07 & 99.85$\pm$0.02 & 99.72$\pm$0.06 \\
\midrule
\multirow{3}{*}{\# heads}
    & 2 &  99.57$\pm$0.03 & 99.37$\pm$0.38 & 98.54$\pm$0.03 & 98.41$\pm$0.05 & 99.63$\pm$0.09 & 99.59$\pm$0.49 \\
    & 4 & 99.58$\pm$0.14 & 99.42$\pm$0.07 & 98.58$\pm$0.02 & 98.44$\pm$0.11 & 99.69$\pm$0.01 & 99.63$\pm$0.03 \\
    & 8 & 99.60$\pm$0.01 & 99.45$\pm$0.03 & 98.63$\pm$0.28 & 98.47$\pm$0.02 & 99.76$\pm$0.02 & 99.66$\pm$0.05 \\
\midrule
\multirow{2}{*}{\# hops}
    & 1 & 99.03$\pm$0.01 & 98.79$\pm$0.62 & 97.65$\pm$0.54 & 97.06$\pm$0.37 & 98.97$\pm$0.05 & 99.02$\pm$0.23 \\
    & 2 & 99.57$\pm$0.03 & 99.37$\pm$0.38 & 98.54$\pm$0.03 & 98.41$\pm$0.05 & 99.63$\pm$0.09 & 99.59$\pm$0.49 \\
\midrule
\multirow{3}{*}{\# 1-hop neighbors}
    & 10 & 99.06$\pm$0.14 & 98.89$\pm$0.02 & 98.49$\pm$0.02 & 98.13$\pm$0.07 & 99.34$\pm$0.21 & 99.12$\pm$0.04 \\
    & 20 & 99.57$\pm$0.03 & 99.37$\pm$0.38 & 98.54$\pm$0.03 & 98.41$\pm$0.05 & 99.63$\pm$0.09 & 99.59$\pm$0.49 \\
    & 30 & 99.23$\pm$0.06 & 99.01$\pm$0.35 & 98.57$\pm$0.03 & 98.36$\pm$0.02 & 99.72$\pm$0.04 & 99.63$\pm$0.05 \\
\midrule
\multirow{3}{*}{\# 2-hop neighbors}
    & 5  & 99.57$\pm$0.03 & 99.37$\pm$0.38 & 98.54$\pm$0.03 & 98.41$\pm$0.05 & 99.63$\pm$0.09 & 99.59$\pm$0.49 \\
    & 10 & 99.65$\pm$0.06 & 99.46$\pm$0.07 & 98.76$\pm$0.01 & 98.95$\pm$0.03 & 99.71$\pm$0.03 & 99.65$\pm$0.02 \\
    & 15 & 99.61$\pm$0.01 & 99.42$\pm$0.05 & 98.89$\pm$0.61 & 99.01$\pm$0.14 & 99.65$\pm$0.04 & 99.63$\pm$0.06 \\
\bottomrule
\end{tabular}
\end{table}

\begin{table}[h!t]
\caption{Hyperparameter analysis in inductive setting}\label{tab:hyperparameter_ind}
\centering
\scriptsize
\renewcommand{\arraystretch}{1.1}
\begin{tabular}{l r | c c c c c c}
\toprule
\multicolumn{2}{c|}{Hyperparameter \& Value}  & \multicolumn{2}{c}{\Wikipedia} & \multicolumn{2}{c}{\UCI} & \multicolumn{2}{c}{\CanParl} \\
 & & AP & AUC-ROC & AP & AUC-ROC & AP & AUC-ROC \\
\midrule
\multirow{3}{*}{\# layers} 
    & 2 & 98.97$\pm$0.25 & 98.97$\pm$0.13 & 98.06$\pm$0.37 & 97.32$\pm$0.54 & 98.12$\pm$0.34 & 98.05$\pm$0.64 \\
    & 4 & 98.84$\pm$0.13 & 98.74$\pm$0.02 & 97.43$\pm$0.81 & 96.83$\pm$0.59 & 98.67$\pm$0.24 & 98.73$\pm$0.01 \\
    & 6 & 98.87$\pm$0.17 & 98.83$\pm$0.29 & 98.17$\pm$0.19 & 97.38$\pm$0.35 & 98.81$\pm$0.04 & 98.96$\pm$0.27 \\
\midrule
\multirow{3}{*}{\# heads}
    & 2 & 98.97$\pm$0.25 & 98.97$\pm$0.13 & 98.06$\pm$0.37 & 97.32$\pm$0.54 & 98.12$\pm$0.34 & 98.05$\pm$0.64 \\
    & 4 & 99.01$\pm$0.16 & 98.99$\pm$0.24 & 98.13$\pm$0.45 & 97.46$\pm$0.19 & 98.36$\pm$0.74 & 98.23$\pm$0.59 \\
    & 8 & 99.03$\pm$0.06 & 99.02$\pm$0.01 & 98.39$\pm$0.05 & 97.59$\pm$0.86 & 98.54$\pm$0.17 & 98.33$\pm$0.32 \\
\midrule
\multirow{2}{*}{\# hops}
    & 1 & 98.45$\pm$0.26 & 98.24$\pm$0.05 & 96.95$\pm$0.70 & 96.37$\pm$0.47 & 97.75$\pm$0.46 & 97.24$\pm$0.54 \\
    & 2 & 98.97$\pm$0.25 & 98.97$\pm$0.13 & 98.06$\pm$0.37 & 97.32$\pm$0.54 & 98.12$\pm$0.34 & 98.05$\pm$0.64 \\
\midrule
\multirow{3}{*}{\# 1-hop neighbors}
    & 10 & 98.67$\pm$0.25 & 98.46$\pm$0.13 & 97.52$\pm$0.15 & 96.79$\pm$0.31 & 98.04$\pm$0.48 & 97.79$\pm$0.12 \\
    & 20 & 98.97$\pm$0.25 & 98.97$\pm$0.13 & 98.06$\pm$0.37 & 97.32$\pm$0.54 & 98.12$\pm$0.34 & 98.05$\pm$0.64 \\
    & 30 & 98.71$\pm$0.46 & 98.52$\pm$0.14 & 98.52$\pm$0.16 & 97.06$\pm$0.27 & 98.65$\pm$0.07 & 98.34$\pm$0.32 \\
\midrule
\multirow{3}{*}{\# 2-hop neighbors}
    & 5 & 98.97$\pm$0.25 & 98.97$\pm$0.13 & 98.06$\pm$0.37 & 97.32$\pm$0.54 & 98.12$\pm$0.34 & 98.05$\pm$0.64 \\
    & 10 & 98.99$\pm$0.05 & 99.01$\pm$0.24 & 97.17$\pm$0.32 & 95.99$\pm$0.22 & 98.75$\pm$0.39 & 98.42$\pm$0.09 \\
    & 15 & 98.93$\pm$0.52 & 98.95$\pm$0.18 & 97.19$\pm$0.38 & 96.02$\pm$0.37 & 98.64$\pm$0.14 & 98.27$\pm$0.59 \\
\bottomrule
\end{tabular}
\end{table}

\subsection{Additional ablation results}\label{app:ablation}

\begin{table*}
\caption{\small Ablation study on 9 datasets in transductive \& inductive settings. }
\label{tab:app:ablation}
\vspace{-4pt}
\centering
\setlength{\tabcolsep}{3pt}
\scriptsize
\textbf{Transductive}\\[0.2em]
\resizebox{0.95\textwidth}{!}{
\begin{tabularx}{\textwidth}{l*{9}{>{\centering\arraybackslash}X}}
\toprule
AUC-ROC & \Wikipedia & \Reddit & \UCI & \Enron & \Mooc & \CanParl & \USLegis & \UNTrade & \UNVote \\
\midrule
\DiffDyG & 99.37$\pm$0.38 & 99.07$\pm$1.24 & 98.41$\pm$0.05 & 98.23$\pm$0.69 & 97.10$\pm$0.27 & 99.59$\pm$0.49 & 81.16$\pm$0.14 & 98.97$\pm$0.34 & 87.49$\pm$1.29 \\
w/o RoPE & 99.03$\pm$1.62 & 99.03$\pm$0.85 & 96.96$\pm$0.01 & 97.19$\pm$4.62 & 95.07$\pm$0.56 & 97.69$\pm$0.26 & 78.03$\pm$0.09 & 95.41$\pm$0.02 & 83.89$\pm$2.14 \\
w/o DA & 98.82$\pm$0.07 & 98.96$\pm$0.34 & 96.22$\pm$0.31 & 93.83$\pm$0.21 & 88.74$\pm$0.81 & 97.78$\pm$0.53 & 77.98$\pm$0.34 & 71.50$\pm$0.23 & 59.17$\pm$0.25 \\
w/o SE & 99.20$\pm$0.30 & 99.05$\pm$0.74 & 97.98$\pm$0.01 & 98.19$\pm$0.61 & 95.26$\pm$0.13 & 98.64$\pm$0.36 & 78.84$\pm$0.68 & 96.05$\pm$0.42 & 85.78$\pm$0.03 \\
\bottomrule
\end{tabularx}
}
\vspace{0.4em}

\textbf{Inductive}\\[0.2em]
\resizebox{0.95\textwidth}{!}{
\begin{tabularx}{\textwidth}{l*{9}{>{\centering\arraybackslash}X}}
\toprule
AP & \Wikipedia & \Reddit & \UCI & \Enron & \Mooc & \CanParl & \USLegis & \UNTrade & \UNVote \\
\midrule
\DiffDyG & 98.97$\pm$0.25 & 99.04$\pm$0.01 & 98.06$\pm$0.37 & 94.48$\pm$1.95 & 95.16$\pm$0.70 & 98.12$\pm$0.34 & 73.78$\pm$0.39 & 97.82$\pm$1.00 & 74.46$\pm$1.82 \\
w/o RoPE & 98.75$\pm$0.26 & 98.73$\pm$0.31 & 96.01$\pm$0.03 & 94.25$\pm$2.79 & 94.66$\pm$0.75 & 96.98$\pm$0.01 & 70.99$\pm$3.51 & 92.11$\pm$0.10 & 66.60$\pm$2.55 \\
w/o DA & 98.45$\pm$0.33 & 98.59$\pm$0.47 & 94.88$\pm$0.51 & 90.74$\pm$0.34 & 87.97$\pm$0.95 & 92.15$\pm$0.24 & 63.09$\pm$0.13 & 65.25$\pm$0.25 & 56.76$\pm$0.54 \\
w/o SE & 98.81$\pm$0.04 & 98.89$\pm$0.56 & 97.73$\pm$1.53 & 93.55$\pm$2.75 & 95.10$\pm$0.25 & 97.57$\pm$0.01 & 70.36$\pm$1.57 & 93.65$\pm$0.13 & 68.21$\pm$1.00 \\
\midrule
AUC-ROC & \Wikipedia & \Reddit & \UCI & \Enron & \Mooc & \CanParl & \USLegis & \UNTrade & \UNVote \\
\midrule
\DiffDyG & 98.97$\pm$0.13 & 98.79$\pm$0.62 & 97.32$\pm$0.54 & 92.59$\pm$0.47 & 95.40$\pm$0.90 & 98.05$\pm$0.64 & 59.38$\pm$1.64 & 98.81$\pm$0.03 & 70.99$\pm$1.16 \\
w/o RoPE & 98.51$\pm$0.03 & 98.47$\pm$2.19 & 95.19$\pm$0.01 & 90.99$\pm$0.33 & 94.27$\pm$1.31 & 96.98$\pm$0.01 & 57.05$\pm$0.60 & 92.08$\pm$1.03 & 66.44$\pm$1.99 \\
w/o DA & 98.48$\pm$0.19 & 98.32$\pm$0.61 & 93.32$\pm$0.26 & 90.47$\pm$0.57 & 87.58$\pm$0.86 & 93.05$\pm$0.67 & 55.26$\pm$0.94 & 68.36$\pm$0.19 & 58.72$\pm$0.49 \\
w/o SE & 98.73$\pm$0.65 & 98.65$\pm$1.34 & 95.60$\pm$1.38 & 91.49$\pm$3.70 & 94.39$\pm$0.93 & 97.37$\pm$0.01 & 56.12$\pm$2.53 & 93.31$\pm$0.52 & 68.69$\pm$0.05 \\
\bottomrule
\end{tabularx}
}
\vspace{-1em}
\end{table*}

Tab.~\ref{tab:app:ablation} provides the ablation studies in transductive setting with AUC-ROC and in inductive settings.

\subsection{Efficiency and model size} 

Fig.~\ref{fig:time_and_para} compares training efficiency and parameter size on Wikipedia and Reddit datasets, where bubble size represents the number of parameters. DiffDyG achieves the best accuracy–efficiency trade-off among all evaluated methods. By leveraging FlashAttention~\cite{dao2022flashattention} and Grouped Query Attention~\cite{ainslie2023gqa}, DiffDyG trains 1.5–2.5$\times$ faster than other Transformer-based models such as DyGFormer and TIDFormer, while maintaining a comparable or smaller model size. This demonstrates that differential attention is not only more effective but also practically efficient for large-scale dynamic graph learning.


\section{The pseudo-code of \textsc{DiffDyG}}

The algorithm of the proposed DiffDyG is summarized in Algo.~\ref{alg:diffdyg}.

\begin{algorithm}[!ht]
\footnotesize
\caption{Training pipeline for \textsc{DiffDyG}}\label{alg:diffdyg}
\begin{algorithmic}[1]
\STATE {\bfseries Input:} Training set $\gG ={(u,v,t, y)}$, number of hop $K$, model with encoder $g$ and classifier $f$
\STATE Initialize model parameters
\FOR{$\text{epoch}=1,2,3,\dots$}
  \FOR{$(u,v,t, y)\in \gG$}
    \FOR{$k=1,\dots,K$}
        \STATE Obtain the $k$-hop neighbor sets of $u$ and $v$ as $\mathcal{H}_k(u,t),\mathcal{H}_k(v,t)$
        \STATE Obtain five feature channels $\mathbf{X}_{k*,N}^t,\mathbf{X}_{k*,E}^t,\mathbf{X}_{k*,T}^t,\mathbf{X}_{k*,C}^t,\mathbf{X}_{k*,S}^t$ with each row corresponding to a node in the $k$-hop neighbor sets. $*$ can be $u$ or $v$.
        \STATE $\mathbf{X}^t_{k*} \gets [\mathbf{X}^t_{k*,N}, \mathbf{X}^t_{k*,E}, \mathbf{X}^t_{k*,T}, \mathbf{X}^t_{k*,C}, \mathbf{X}^t_{k*,S}] $
    \ENDFOR
    \STATE Obtain multi-hop encoding representations: $\mathbf{Z}_{*}^t \gets [\mathbf{X}_{1*}^t, \cdots,  \mathbf{X}_{K*}^t]$
    \STATE Obtain final representations $\mathbf{Y}_{*}^t=g(\mathbf{Z}_{*}^t)$
    \STATE Compute the link probability $p \gets f([\mathbf{Y}_u^t, \mathbf{Y}_v^t])$
    \STATE Compute the loss $\mathcal{L}_{BCE}(p, y)$
    \STATE Update the model
\ENDFOR
\ENDFOR
\end{algorithmic}
\end{algorithm}

\newpage

\end{document}